\documentclass[pdflatex,sn-basic]{sn-jnl}% Basic Springer Nature Reference Style
\usepackage{graphicx}%
\usepackage{multirow}%
\usepackage{amsmath,amssymb,amsfonts}%
\usepackage{amsthm}%
\usepackage{mathrsfs}%
\usepackage[title]{appendix}%
\usepackage{xcolor}%
\usepackage{textcomp}%
\usepackage{manyfoot}%
\usepackage{booktabs}%
\usepackage{algorithm}%
\usepackage{algorithmicx}%
\usepackage{algpseudocode}%
\usepackage{listings}%
\usepackage{etoolbox}%
\usepackage{xurl}%
\hypersetup{hypertexnames=false}

\theoremstyle{thmstyleone}%
\theoremstyle{thmstyletwo}%

\theoremstyle{thmstylethree}%

\makeatletter
\AtBeginDocument{%
  \renewcommand{\bibfont}{%
    \reset@font\fontfamily{\rmdefault}\normalsize\selectfont
    \rightskip=0pt plus 2em\relax
    \spaceskip=0pt\relax
    \xspaceskip=0pt\relax
    \emergencystretch=2em\relax
  }%
}
\makeatother
\begin{document}

\title[LandslideAgent]{LandslideAgent with Multimodal LandslideBench: A Domain-Rule-Augmented Agent for Autonomous Landslide Identification and Analysis}


\author[1]{\fnm{Chengfu} \sur{Liu}}

\author*[1]{\fnm{Dongyang} \sur{Hou}}\email{houdongyang1986@csu.edu.cn}

\author[1]{\fnm{Junwu} \sur{Xiang}}

\author[1]{\fnm{Cheng} \sur{Yang}}

\author[1]{\fnm{Xuezhi} \sur{Cui}}

\author[1]{\fnm{Zeyuan} \sur{Wang}}

\author[1]{\fnm{Liangtian} \sur{Liu}}

\author[1]{\fnm{Zelang} \sur{Miao}}

\affil[1]{\orgdiv{School of Geosciences and Info-Physics}, \orgname{Central South University}, \orgaddress{\street{932 South Lushan Road}, \city{Changsha}, \postcode{410083}, \state{Hunan}, \country{China}}}


\abstract{Intelligent landslide hazard interpretation is critical for disaster prevention, yet current paradigms struggle to simultaneously extract visual features and high-level geoscientific semantics, while general-purpose vision–language models (VLMs) suffer from perceptual limitations and domain hallucinations in complex geological scenarios. To address these challenges, we propose an instruction-driven agentic framework comprising three components. First, LandslideBench, a multimodal fine-grained dataset with seven subtype labels, high-resolution imagery, pixel-level masks, and high-quality textual descriptions, is constructed via multi-VLM cross-validation and interactive annotation. Then, LandslideVLM, a landslide-oriented VLM, is fine-tuned via LoRA on LandslideBench to enhance geological semantic understanding. Finally, LandslideAgent, a domain rule-enhanced agent taking LandslideVLM as its cognitive backbone, employs a dual-rule controller incorporating structured report metadata constraints and cross-validation identification constraints to regulate automated tool invocation. Experiments demonstrate that LandslideBench provides effective baselines across five mainstream models on fine-grained classification and semantic segmentation. LandslideVLM achieves accuracy improvements of 10.96\%, 32.87\%, and 15.91\% on landslide discrimination, fine-grained classification, and semantic description quality, respectively. LandslideAgent further enables autonomous multi-source spatial data inference, realizing full-process intelligence for landslide identification and analysis.}




\maketitle

\section{Introduction}\label{sec1}
Landslides are globally pervasive and highly destructive geological 
hazards that pose severe threats to human life, critical infrastructure, 
and ecological stability~\citep{yunjian20262025}. Consequently, the rapid and precise 
identification of landslide hazards is essential for effective risk 
assessment, emergency response, and post-disaster recovery~\citep{fang_rapid_2026}. However, 
traditional field surveys are inherently time-consuming and 
labor-intensive; moreover, they present significant safety risks and 
operational challenges in inaccessible or steep terrains. Driven by 
recent advances in Earth observation, remote sensing-based approaches 
have become indispensable for landslide identification and analysis, 
offering unparalleled advantages in spatial coverage, multiscale 
monitoring, high temporal resolution, and operational safety~\citep{casagli2023landslide}.

Currently, remote sensing-based landslide identification is undergoing a 
paradigm shift from traditional visual interpretation to purely 
vision-based deep learning approaches, predominantly driven by 
Convolutional Neural Networks (CNNs), Vision Transformers (ViTs) and so 
on~\citep{fu2026cnn,zhang_analysis_2024}. Through object detection or semantic segmentation, these methods 
extract deep features from historical annotated inventories to delineate 
landslide bounding boxes or precise contours~\citep{yang_feature_2025}. For example, recent 
advancements have seen the integration of efficient backbones and 
attention mechanisms into lightweight detection models, as well as 
the use of wavelet-guided modules to enhance boundary sensitivity in 
semantic segmentation~\citep{song2025landslide,zhou2026landslide}. Despite these architectural improvements, most 
methods remain confined to visual perception of landslides, focusing 
solely on spatial localization of landslides while failing to analyze 
the environmental context and triggering factors of landslides. This 
limitation arises from two main aspects: first, existing landslide 
datasets primarily provide binary semantic labels (landslide vs. 
non-landslide) along with boundary masks or extents, lacking richer 
semantic metadata; second, pure-vision models inherently lack the 
capacity for cross-modal reasoning and the integration of external 
domain knowledge. As a result, current landslide identification 
fundamentally remains a spatial recognition task, leaving comprehensive 
hazard analysis highly dependent on manual empirical judgment.

To overcome the inherent limitations of pure-vision models in 
high-level cognition and reasoning, Vision-Language Models (VLMs) have 
recently been introduced, offering a novel end-to-end paradigm for 
landslide analysis. For instance, an inspiring pioneering study by 
Areerob et al. leveraged Multimodal Large Language Models (MLLMs) to 
translate structured domain expertise into natural language reports~\citep{areerob2025multimodal}. 
Their approach directly generates comprehensive insights---such as disaster 
types, causal factors, and risk predictions---from single images of affected 
areas. While their work lays a crucial foundation for high-level disaster 
cognition, the broader application of the current VLM paradigm faces 
inherent challenges stemming from the ``black-box'' nature of direct 
vision-language alignment. Specifically, current end-to-end architectures 
generally encounter difficulties in seamlessly integrating existing 
high-precision, domain-specific visual tools (e.g., the aforementioned 
pure-vision deep learning models for object detection and segmentation). 
Consequently, relying solely on these models to delineate 
precise landslide boundaries or discern fine-grained local textures 
remains a significant challenge. Furthermore, the development of such 
models is heavily reliant on multimodal alignment data, which is 
exceptionally difficult to acquire in the domain of geological 
hazards. For example, the exploratory dataset utilized in the 
aforementioned study comprised 68 landslide images, highlighting the 
pervasive issue of data scarcity in this nascent field. This limited 
availability of large-scale domain-specific data inevitably constrains 
the generalization capabilities of VLMs across complex, real-world 
environments.

These shortcomings suggest that many current VLM solutions remain constrained by an end-to-end, 
single-model formulation, in which visual perception and 
reasoning are often entangled, thereby limiting their ability to 
support dynamic, modular, and multi-tool collaboration~\citep{Suris_2023_ICCV}.
 Therefore, to further elevate the intelligence and reliability of landslide analysis, there is 
 an urgent need to break away from the framework of single end-to-end 
 models. This requires not only the construction of larger-scale, 
 fine-grained multimodal landslide datasets to fine-tune and strengthen 
 the foundational cognition of large models, but also the urgent 
 exploration of novel architectures equipped with autonomous planning 
 and tool-invocation capabilities. By seamlessly orchestrating multiple 
 mature, high-precision specialized tools, such an architecture can 
 effectively bridge the gap between multi-dimensional feature perception 
 and causal reasoning, ultimately achieving a comprehensive and highly 
 interpretable analysis of landslide hazards.

 In fact, the autonomous agent paradigm offers a transformative 
 perspective for overcoming the limitations inherent in traditional 
 end-to-end models~\citep{wang_survey_2024,tang2025intelligent}. Typically centered on large language models (LLMs) 
 or multimodal foundation models as the core cognitive engine, 
 intelligent agents possess capabilities in task understanding, multi-step 
 planning, external tool invocation, and state management. Unlike 
 conventional models that rigidly map remote sensing imagery directly to 
 outputs (e.g., classes, spatial masks, or captions), agents can 
 dynamically decompose complex tasks into executable sub-steps, adaptively 
 invoking specialized models and external knowledge bases during inference. 
 This mechanism enables a seamless synergy between general-purpose 
 semantic reasoning and domain-specific computational processes. 
 Recently, agent-based approaches have demonstrated remarkable 
 performance across Earth observation~\citep{talemi2026agenticairemotesensing} and disaster management domains~\citep{chen2026integration}. 
 For instance, RS-Agent~\citep{xu2024rs} employs a large-model-based central controller 
 combined with a dynamic toolset to automate multi-scenario remote 
 sensing tasks while Earth-Agent~\citep{feng2025earth} integrates hundreds of Earth 
 observation-specific tools to support multimodal quantitative 
 spatiotemporal reasoning. More specifically in the realm of hazard 
 management, pioneering frameworks such as DisasterReliefGPT~\citep{reghunath2026disasterreliefgpt} have 
 successfully utilized multimodal agents for autonomous disaster 
 impact assessment and crisis communication. These studies collectively 
 underscore the immense potential of the agent paradigm in advancing 
 intelligent hazard analysis. 

However, landslide analysis inherently demands strict logical hierarchies 
and strong evidentiary dependencies, as dictated by rigorous geoscientific 
protocols~\citep{feng_convolutional_2025,alcantara-ayala_landslides_2025}. Although general-purpose agents excel at dynamic task 
orchestration, their lack of explicit domain constraints often leads 
them to prioritize immediate task completion. Consequently, they are 
prone to bypassing critical intermediate evidence-gathering steps---such 
as precise mask extraction or contextual terrain comparison---resulting in 
broken, uninterpretable, or hallucinated reasoning chains. Therefore, it 
is imperative to design a domain-specialized agent embedded with 
landslide-specific analytical rules. Such an architecture is crucial to 
guaranteeing the rigor of the reasoning process and the reliability of 
analytical outcomes in complex geohazard interpretation tasks.

To address the aforementioned challenges, this paper proposes an 
instruction-driven agentic framework for landslide disaster response, 
transitioning from multimodal perception to the generation of specialized 
analytical reports. This architecture is grounded in a newly constructed 
multimodal benchmark and driven by a domain rule-augmented agent. 
Specifically, we first construct LandslideBench, a fine-grained multimodal 
dataset that intricately integrates remote sensing imagery, precise 
pixel-level masks, and rich semantic textual descriptions, thereby 
establishing a robust data foundation for domain-specific knowledge 
transfer. Leveraging this foundation, LandslideVLM, a vision-language 
foundation model, is developed through domain-adaptive fine-tuning to 
comprehend complex geoscientific scenes and generate domain semantic 
content. Finally, using LandslideVLM as the central cognitive engine, we 
propose LandslideAgent---a domain-rule-augmented landslide agent---to address the 
logical inconsistencies that general-purpose agents encounter in complex 
reasoning tasks.

(1) The construction of LandslideBench, a fine-grained multimodal dataset 
tailored for landslide analysis. Extending beyond conventional spatial 
masks, this dataset incorporates seven distinct landslide subtype labels, 
non-landslide negative samples, and comprehensive textual 
descriptions (e.g., morphological and environmental characteristics). 
Encompassing 2,130 precise image-text pairs, it provides a rich 
structural foundation for semantic-level analysis and cross-modal 
learning in geosciences.

(2) The development of LandslideVLM, a domain-specific vision-language 
model for enhanced geoscientific reasoning. Fine-tuned on the proposed 
benchmark, this model extends standard visual perception capabilities. 
It enables the simultaneous execution of precise landslide existence 
judgment and the generation of analytical descriptions that are rigorously 
aligned with professional geoscientific contexts, thereby elevating the 
model's interpretive depth in complex terrains.

(3) The design of LandslideAgent, an intelligent system governed by 
dual-rule constraints for landslide identification 
and analysis. Through natural language interaction, the agent 
autonomously orchestrates professional tools, including semantic 
segmentation and terrain-geological information retrieval. 
By systematically integrating domain rules, it offers a highly 
practical paradigm that shifts landslide response from isolated 
perceptual tasks toward cognition-oriented, intelligent analysis. 
\section{Method}\label{sec2}
\begin{figure}[htbp]
\centering
\includegraphics[width=\textwidth]{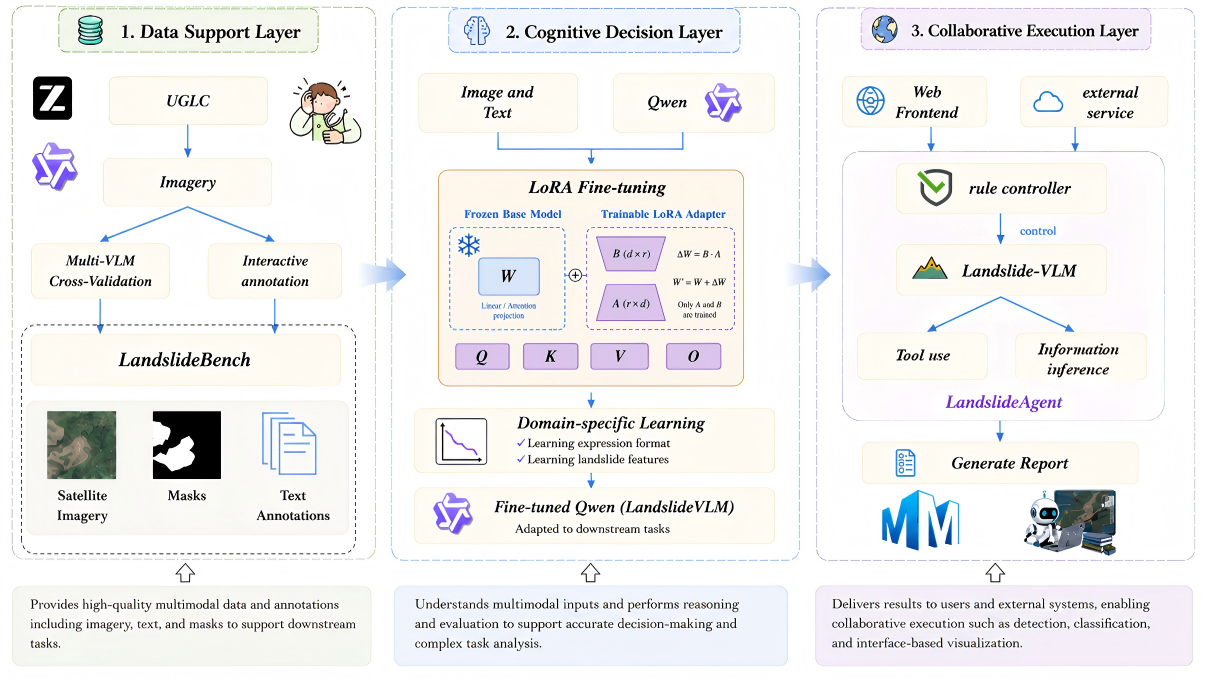}
\caption{Technical roadmap of this study: dataset construction, model fine-tuning, and agent construction}
\label{fig:techroad}
\end{figure}

To transcend the perceptual constraints of conventional single-modality 
models in complex geological scenarios, we propose an integrated, 
hierarchical agentic architecture for automated landslide interpretation 
and analysis. As illustrated in Figure~\ref{fig:techroad}, the proposed framework is 
systematically orchestrated across three interdependent layers: Data 
Support, Cognitive Decision-Making, and Collaborative Execution. At the 
Data Support Layer, we establish a standardized multimodal pipeline to 
construct LandslideBench. By extracting and refining specific landslide 
subtypes from the Unified Global Landslide Catalogue (UGLC)~\citep{mancino2025unified}, this layer rigorously aligns 
high-resolution imagery, pixel-level spatial masks, and structured 
textual descriptions, employing a human-in-the-loop strategy and a 
dual-model verification mechanism. This alignment transforms raw 
heterogeneous data into a cohesive structural foundation, supplying 
indispensable multi-dimensional inputs for downstream domain adaptation 
and logical reasoning. At the Cognitive Decision-Making Layer, the 
framework develops LandslideVLM through the domain-adaptive fine-tuning 
of the foundational Qwen3-VL-8B-Instruct~\citep{bai2025qwen3}. Functioning as the 
central cognitive engine (``brain'') of the system, this specialized 
model translates visual representations into high-level geoscientific 
semantics, specifically empowering the intricate interpretation of 
landslide triggers, environmental contexts, and morphological attributes. 
At the Collaborative Execution Layer, the proposed 
LandslideAgent integrates LandslideVLM with specialized 
visual models and external geospatial tools under 
domain-rule constraints, enabling cross-validation, 
spatial reasoning, and structured report generation. 
Through the coordination of these three layers, the 
proposed architecture systematically converts 
heterogeneous remote sensing inputs into actionable 
geoscientific insights and comprehensive analytical 
reports.

\subsection{Data Support Layer: Curating the Multimodal LandslideBench}\label{subsec:data-support}
\begin{figure}[htbp]
\centering
\includegraphics[width=\textwidth]{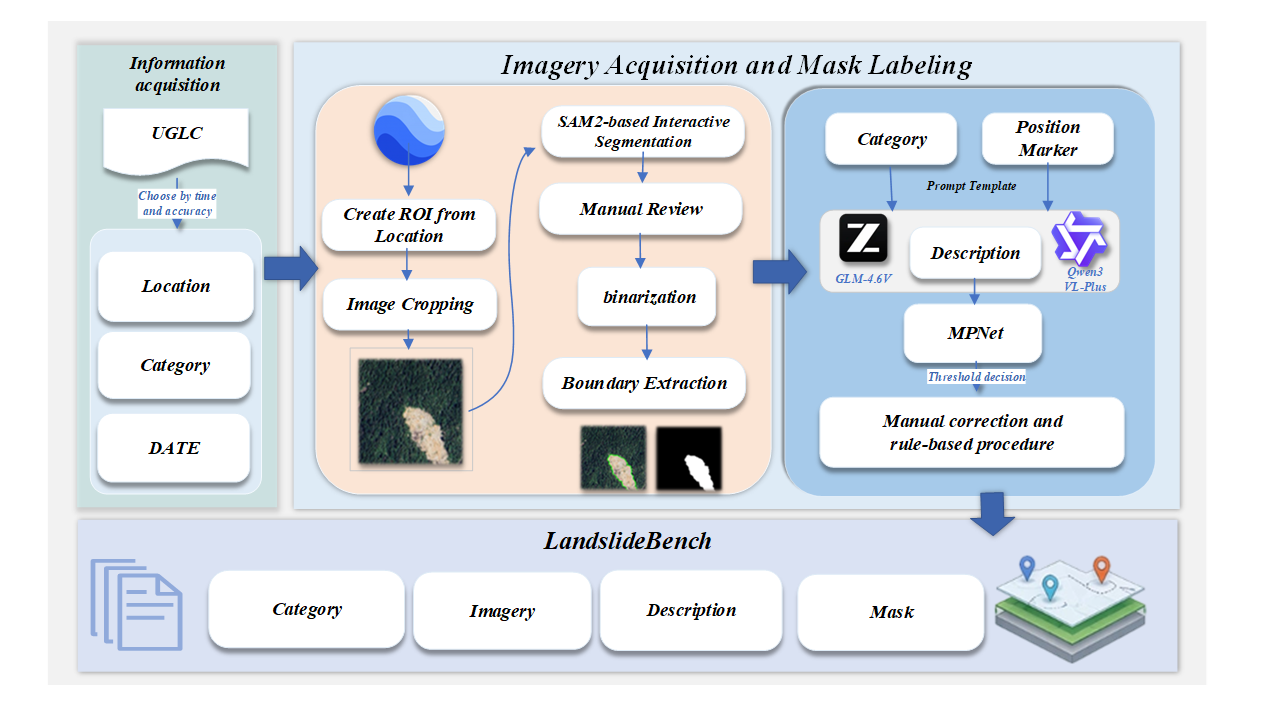}
\caption{Overview of the LandslideBench dataset construction workflow}
\label{fig:dataset-construction}
\end{figure}

While prominent remote sensing-based landslide 
datasets (e.g., Landslide4Sense~\citep{ghorbanzadeh2022landslide4sense}, HR-GLDD~\citep{meena2023hr}, the Bijie dataset~\citep{ji2020landslide}, LMHLD~\citep{liu2025lmhld}, and 
CAS~\citep{xu2024cas}) have advanced automated detection, they are predominantly 
constrained to binary segmentation tasks. Consequently, they inherently 
lack fine-grained sub-type categorization and multi-dimensional semantic 
annotations. To bridge this critical gap in deep semantic representation, 
we curate LandslideBench, a comprehensive multimodal dataset explicitly 
engineered to encapsulate multidimensional geoscientific semantics. The 
systematic workflow for its curation is illustrated in Figure~\ref{fig:dataset-construction}.

To ensure spatial precision and strict semantic alignment, we initiate 
the curation process by extracting baseline metadata---specifically spatial 
coordinates, temporal records, and landslide subtypes---from the UGLC. Using these geospatial anchors, we 
acquire corresponding high-resolution remote sensing imagery. For 
spatial annotation, we implement a human-in-the-loop protocol to 
accurately delineate the pixel-level boundaries of the landslide bodies. 
Concurrently, the textual generation pipeline is governed by a tripartite 
verification mechanism: dual-model rule constraints, automated 
cross-validation, and rigorous expert review. This ensures the 
production of structured scene descriptions across nine specialized 
geoscientific dimensions, including morphological attributes, kinematic 
mechanisms, and triggering factors. Ultimately, we standardize these 
heterogeneous data streams via a unified template, yielding strictly 
aligned triplet samples (image, pixel-mask, and natural language text) 
to support downstream multimodal joint representation learning.
\subsubsection{Landslide Taxonomy and Geospatial Metadata Extraction}\label{subsubsec:taxonomy-metadata}
We base our categorical framework on the established Cruden and Varnes 
classification matrix~\citep{hungr2014varnes}, which is inherently embedded within the UGLC. 
This taxonomic system systematically intersects movement 
materials (e.g., rock, debris, earth) with kinematic 
mechanisms (e.g., fall, slide, flow) to precisely define the physical 
failure modes.

In order to construct a dataset that optimally balances morphological 
diversity with annotation feasibility, seven predominant subtypes 
have been deliberately isolated from this taxonomy as core research 
targets: rock fall, rock slide, earth slide, debris flow, earth flow, 
mud flow, and mudslide. These specific typologies dominate global 
incidence records, thereby ensuring broad representativeness. 
Subsequent to this taxonomic selection, the associated geospatial 
metadata is extracted from the UGLC. In order to guarantee data 
fidelity and strict spatial accuracy, a rigorous filtration process 
is applied to the inventory, retaining only those selected events 
possessing a ``Reliability'' index between 1 and 3. These highly credible 
coordinates subsequently serve as foundational spatial anchors for 
acquiring high-resolution remote sensing imagery.

Finally, in order to mitigate the occurrence of false-positive detections 
in practical inference scenarios, a calibrated proportion of non-landslide 
background samples is injected into the system. These negative anchors 
act as crucial constraints, enhancing the discriminative robustness of 
the multimodal representation learning.

\subsubsection{Multimodal Triplet Construction and Semantic Alignment Pipeline}\label{subsubsec:triplet-construction}
To facilitate downstream multimodal representation learning, we construct 
strictly aligned Image-Mask-Text triplets anchored to the previously 
extracted geospatial metadata.

\begin{itemize}
  \setlength{\leftskip}{1em}
  \item \textbf{Image Acquisition:} Utilizing the filtered UGLC coordinates as spatial 
  centroids, we define precise Regions of Interest (ROIs) to extract 
  true-color (RGB) imagery from the Google Earth platform. Following 
  rigorous quality screening, we compile a curated subset of 2,130 valid 
  samples. These images, primarily sourced from zoom levels 15--16, offer 
  a spatial resolution of 2.4 to 4.8 m/pixel. This specific scale 
  optimally resolves the morphological boundaries and textural details 
  of the hazard bodies, establishing a reliable foundation for subsequent 
  pixel-level annotation.
  \item \textbf{Human-in-the-loop Mask Annotation:} To generate high-fidelity spatial masks, we implement an interactive, human-in-the-loop 
  segmentation paradigm. Leveraging the Segment Anything Model 2 (SAM 2)~\citep{ravi2024sam}
  integrated within the Labelme framework, annotators first generate 
  baseline landslide contours via point/box prompts. Subsequently, 
  experts meticulously refine local mis-segmentations and boundary 
  artifacts, ensuring strict morphological accuracy while drastically 
  accelerating the annotation workflow.
  \item \textbf{Structured Text Generation and Validation:} For the semantic modality, we devise a structured, multi-dimensional generation strategy to ensure geoscientific comprehensiveness. 
  We input prior UGLC subtype labels, the raw imagery, and mask-derived 
  bounding contours as unified multimodal prompts. These guide the 
  Vision-Language Models (VLMs) to synthesize scene descriptions across 
  nine core dimensions, including landslide type, relative position 
  within the image, morphological characteristics, material composition, 
  movement features, environmental context, impacts on human 
  infrastructure, criteria for fine-grained classification, and 
  causal inference. Crucially, to mitigate the inherent hallucination 
  risks of VLMs, we deploy a robust dual-model parallel inference 
  framework (utilizing GLM-4.6V~\citep{hong2025glm} and Qwen3-VL-Plus). To counter the 
  common spatial reasoning drift in visual models, we extract 
  deterministic relative position coordinates via hard-coded rules, 
  injecting them as hard constraints into the prompt template. 
  Subsequently, an MPNet-based~\citep{song2020mpnet} sentence embedding model evaluates the 
  semantic concordance between the dual outputs. Text pairs exhibiting 
  a cosine similarity below 0.8 are flagged for targeted inspection to 
  filter out typical visual hallucinations (e.g., phantom infrastructure 
  or misclassified debris). Ultimately, all nine-dimensional descriptions 
  undergo rigorous manual auditing by domain experts. This stringent 
  pipeline yields multimodal triplets that are syntactically standardized, 
  semantically dense, and rigorously aligned with geoscientific reality.
\end{itemize}

\subsubsection{LandslideBench Overview and Multimodal Statistics}\label{subsubsec:landslidebench-statistics}
\begin{figure}[htbp]
\centering
\includegraphics[width=\textwidth]{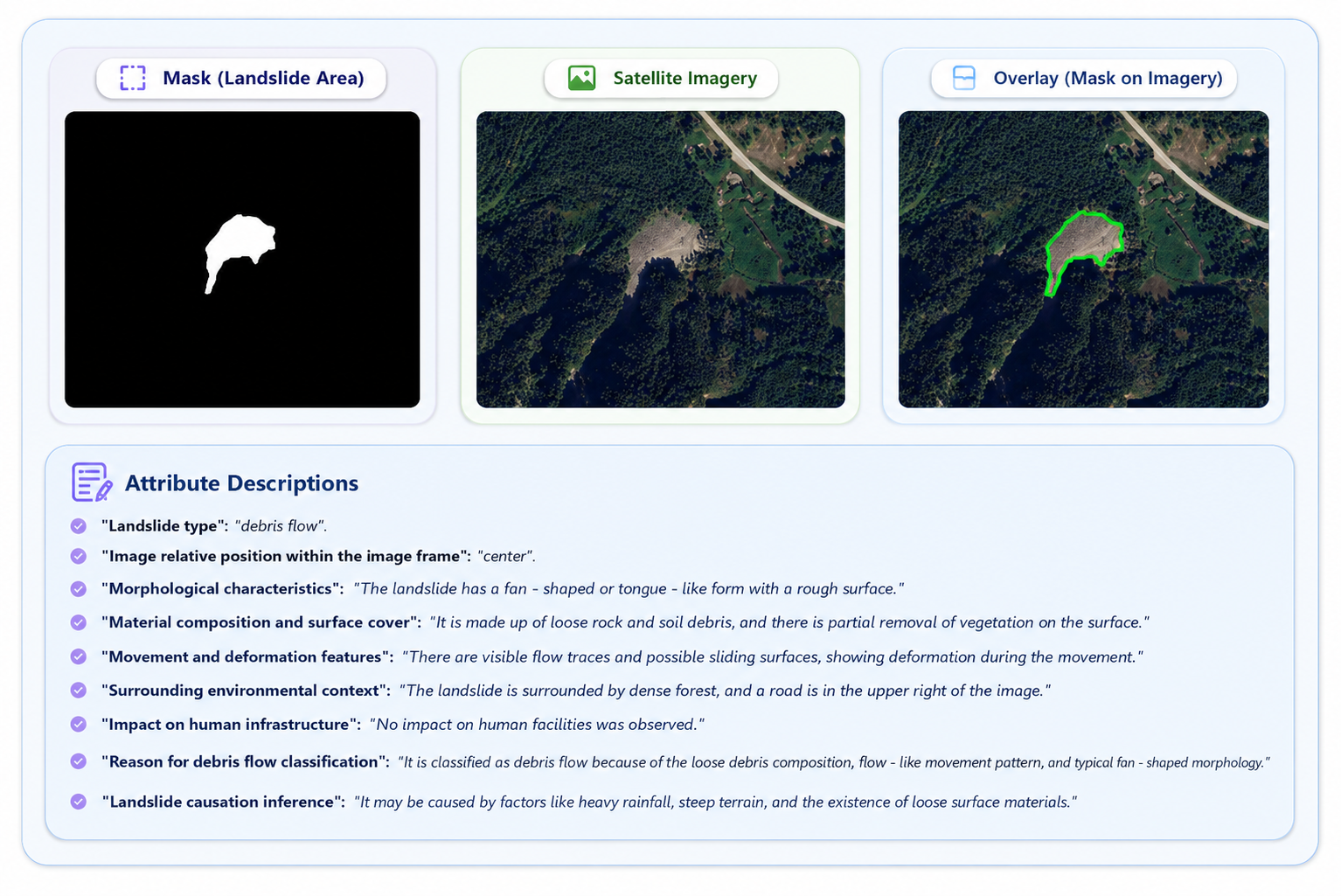}
\caption{Representative examples from the LandslideBench dataset, illustrating the aligned image-mask-text triplets constructed for multimodal landslide understanding}
\label{fig:dataset-instance}
\end{figure}

Overall, LandslideBench comprises 2,130 samples. Each sample strictly 
adheres to the aligned Image-Mask-Text triplet 
structure (as illustrated in Figure~\ref{fig:dataset-instance}). This architectural design 
establishes a definitive correspondence among raw visual signals, 
explicit spatial constraints, and dense semantic concepts, thereby 
enabling simultaneous support for conventional vision tasks and advanced 
multimodal reasoning.

Geographically, the dataset spans 16 countries across Asia, Europe, the 
Americas, and Africa. At the macroscopic scene level, the imagery 
encapsulates a broad spectrum of land-cover types, ranging from densely 
forested mountains to bare terrain and anthropogenically modified regions. 
This inherent environmental heterogeneity ensures that the hazard 
morphologies are heavily modulated by diverse background conditions, 
significantly increasing the scene complexity.

\begin{figure}[htbp]
\centering
\includegraphics[width=\textwidth]{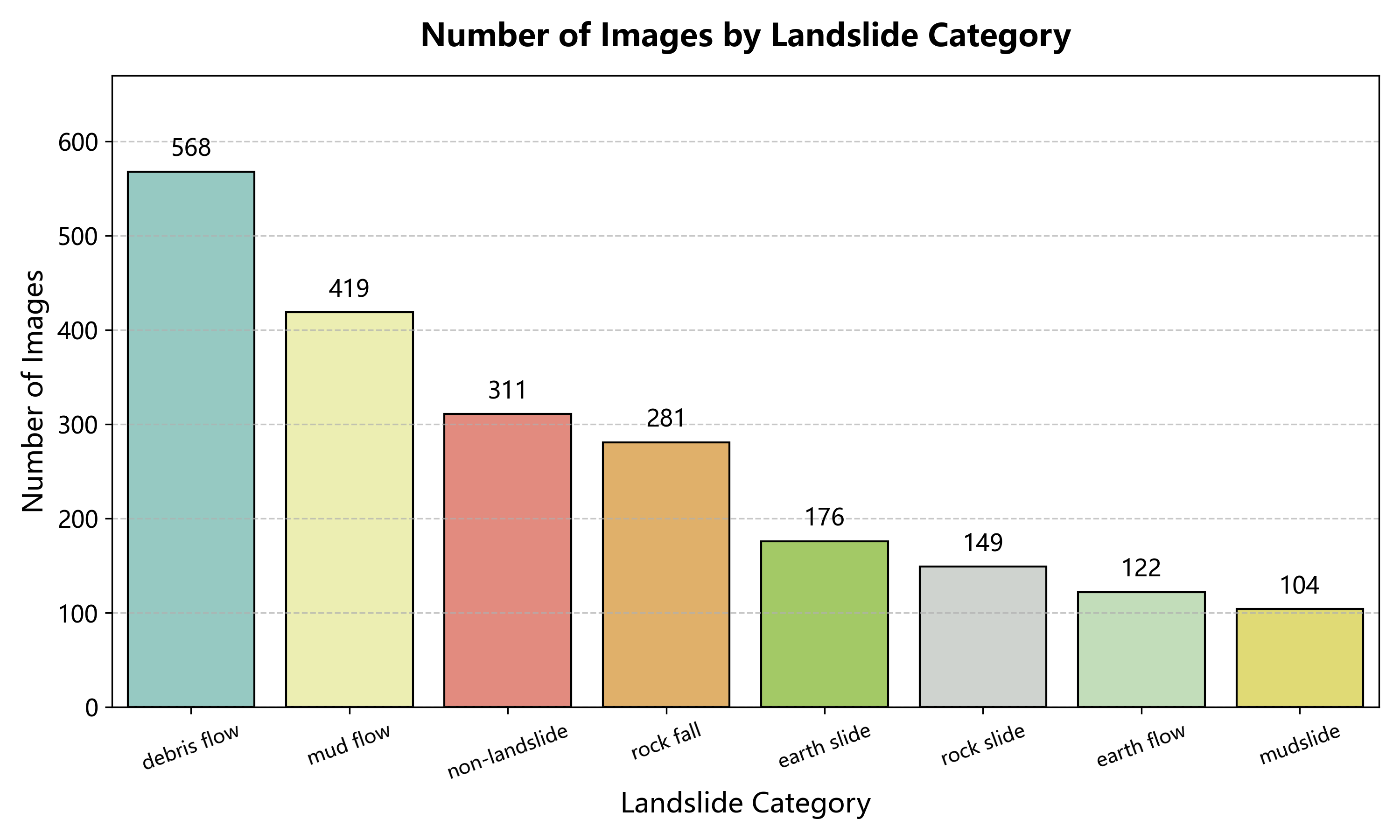}
\caption{The sample size of landslide types}
\label{fig:dataset-size}
\end{figure}

Taxonomically, the dataset naturally exhibits a long-tailed distribution 
across the seven subtypes (Figure~\ref{fig:dataset-size}). Rather than artificially enforcing 
class balance through synthetic resampling, we deliberately preserve this 
authentic class distribution to mirror the true prior probability of 
global landslide occurrences. Consequently, the dataset provides a 
highly realistic benchmark for evaluating model robustness under 
real-world class imbalances.

\begin{figure}[htbp]
\centering
\begin{minipage}[t]{0.48\textwidth}
  \centering
  \includegraphics[width=\linewidth,height=5cm,keepaspectratio]{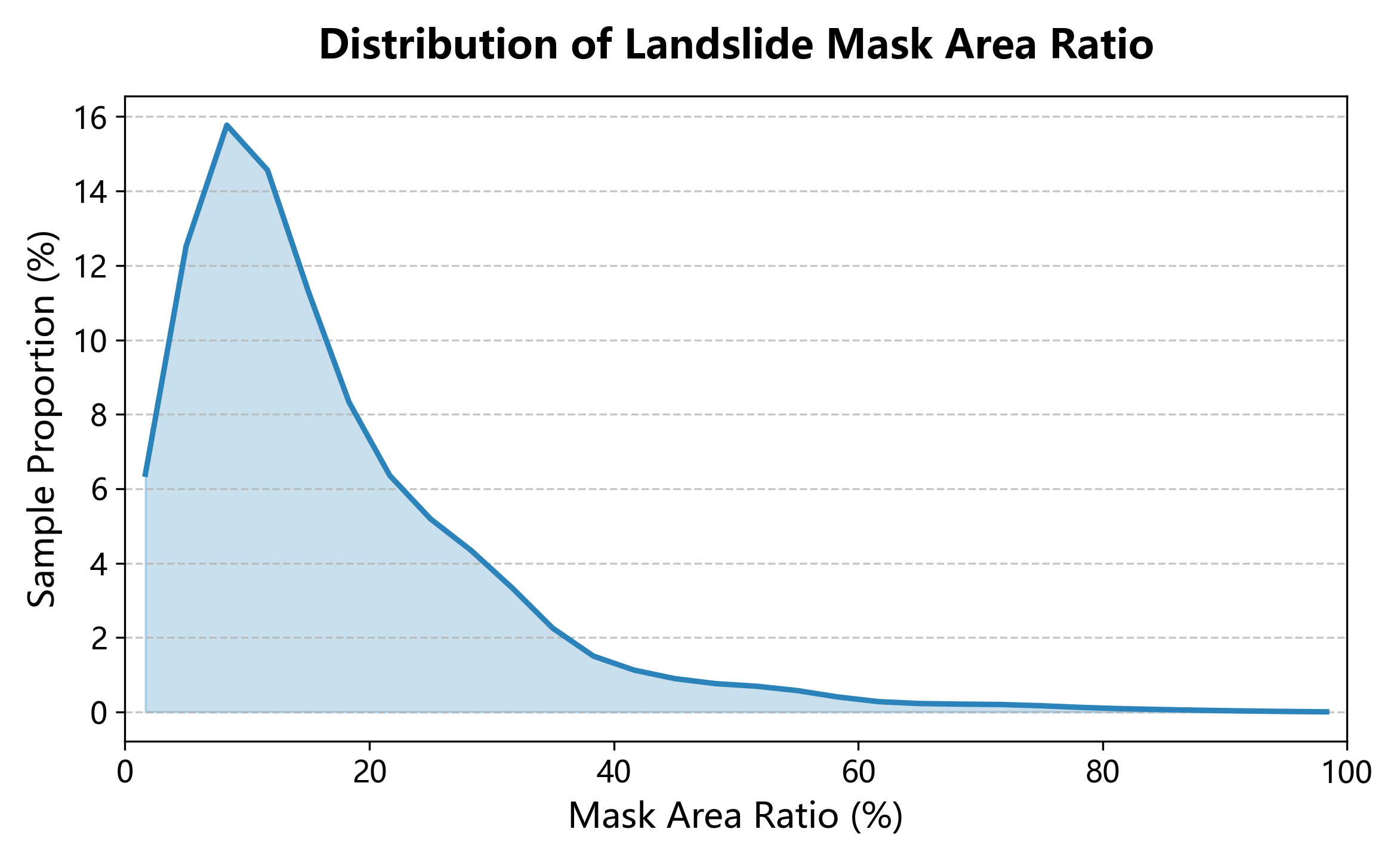}
  \caption{Area-ratio distribution of landslide instances in LandslideBench}
  \label{fig:square-ratio}
\end{minipage}
\hfill
\begin{minipage}[t]{0.48\textwidth}
  \centering
  \includegraphics[width=\linewidth,height=5cm,keepaspectratio]{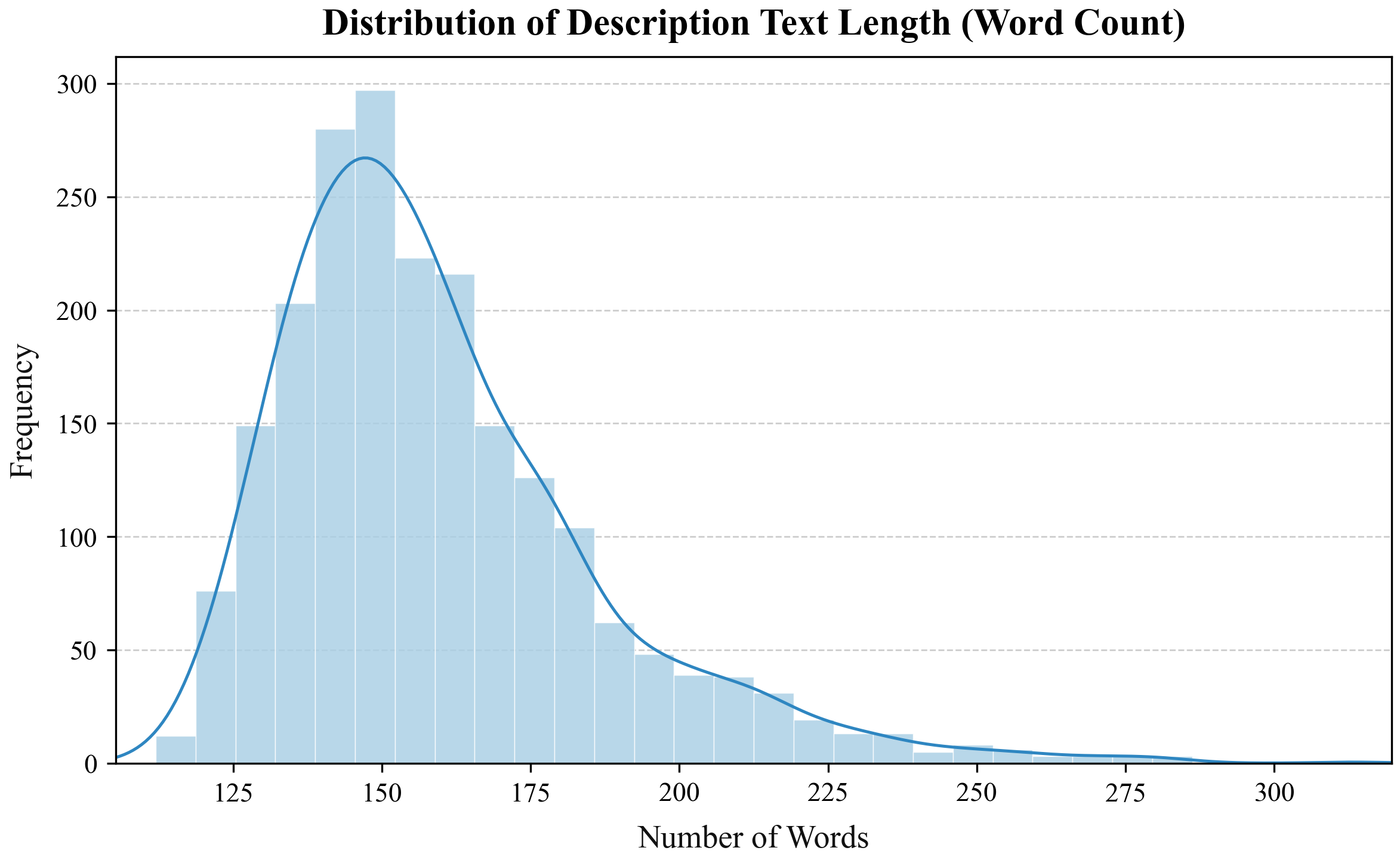}
  \caption{Word-count distribution of textual descriptions in LandslideBench}
  \label{fig:vocabulary-count}
\end{minipage}
\end{figure}

At the instance level, the target hazards demonstrate severe 
morphological variance, manifesting as elongated, blocky, or 
irregularly diffusive patterns. Statistical analysis reveals that 
the target-to-image area ratio spans dramatically from 
0.19\% to 86.97\% (mean: 13.14\%, median: 8.89\%). This massive 
intra-class scale variation (Figure~\ref{fig:square-ratio}) establishes a rigorous testbed 
for scale-invariant representation learning.

Regarding the semantic modality, the descriptive texts range from 112 to 
313 words per sample. More importantly, the length distribution is highly 
concentrated (mean: 159.4, median: 154, standard deviation: 26.8), as 
depicted in Figure~\ref{fig:vocabulary-count}. This low structural variance guarantees high semantic 
density and expressive consistency across the dataset, which is crucial 
for mitigating text-length bias during vision-language contrastive 
optimization.

\subsection{Cognitive Decision-Making Layer: Fine-Tuning the LandslideVLM}\label{subsec:landslidevlm-finetuning}
While general-purpose VLMs possess foundational multimodal reasoning 
capabilities, their direct application to remote sensing is heavily 
constrained by the legacy of single-modal datasets. Consequently, 
off-the-shelf VLMs lack deep geoscientific semantic grounding. First, 
the absence of bidirectional mapping between visual features and expert 
knowledge leads to severe textural ambiguity. For instance, general models 
frequently confound landslide deposits with riverbank erosion or 
bare soil, failing to achieve unbiased generalization across complex 
geological contexts.

Furthermore, landslide interpretation extends beyond mere spatial 
localization; it is a composite cognitive task demanding causal inference 
and fine-grained taxonomy. Without domain-specific interpretative corpora, 
VLMs are prone to epistemic hallucinations---prioritizing linguistic 
fluency over geological rigor and fabricating spurious logical 
connections. Because such hallucinations inevitably propagate errors 
into downstream tool invocation and report synthesis, zero-shot prompt 
engineering is fundamentally insufficient. Therefore, domain-specific 
adaptation driven by specialized datasets is imperative to establish 
robust visual-semantic alignments tailored for earth sciences.

To instantiate the cognitive core of our agent framework, we adopt 
Qwen3-VL-8B-Instruct as the backbone VLM, owing to its robust joint 
vision-text processing capabilities. To bridge the domain gap under 
constrained computational budgets, we perform parameter-efficient 
fine-tuning (PEFT) via Low-Rank Adaptation (LoRA)~\citep{hu2022lora} within the 
LlamaFactory~\citep{zheng2024llamafactory} framework, yielding the domain-expert model, LandslideVLM. 
Rather than updating the entire parameter space, this strategy freezes 
the pre-trained weights while incrementally optimizing the attention 
mechanisms. This process explicitly injects the spatial priors and 
geoscientific semantics derived from LandslideBench into the model, 
effectively suppressing domain-specific hallucinations and textural 
misclassifications.

Specifically, the LoRA modules are integrated into the 
Query (Q), Key (K), Value (V), and Output (O) projection matrices 
within the attention blocks. Given an input hidden state $X \in \mathbb{R}^{n \times d}$, the
modified projection operation is formulated as:
\begin{equation}
Q = XW_Q,\quad K = XW_K,\quad V = XW_V
\end{equation}

\begin{equation}
O = \mathrm{Attention}(Q, K, V)W_O
\end{equation}

Where $W_Q$, $W_K$, $W_V$, $W_O$ denote the projection matrices of the Query, Key,
Value, and Output in the attention mechanism, respectively. In this 
work, a LoRA-based low-rank update is introduced to the above projection 
matrices $W_* \in \{W_Q, W_K, W_V, W_O\}$, and the update formulation can be expressed as:
\begin{equation}
W_*' = W_* + \frac{\alpha}{r} B_* A_*,
\quad * \in \{Q, K, V, O\}
\end{equation}

Where $A_* \in \mathbb{R}^{r \times d}$, $B_* \in \mathbb{R}^{d \times r}$. The low-rank dimension ($r$) is set to 8, and the scaling
factor ($\alpha$) is set to 16 to regulate the contribution of the injected
knowledge. During fine-tuning, the original parameters ($W_*$) of the
backbone model are frozen, and only the low-rank matrices ($A_*$) and ($B_*$)
are optimized to capture domain-specific features, thereby enabling 
knowledge transfer with minimal computational overhead.

During the optimization phase, the fine-tuning is executed over 5 epochs with an effective batch size of 8. To ensure training stability, the 
initial learning rate is set to $2\times10^{-4}$, strictly regulated by 
a cosine annealing schedule following a 5\% linear warm-up phase.

Through this domain-specific fine-tuning, the proposed LandslideVLM 
is able to more effectively learn the semantic representations of 
landslide-related features in remote sensing imagery, including texture, 
boundaries, morphology, and surrounding environmental context. 
Consequently, the model establishes a robust cross-modal alignment 
between raw visual signals and structured analytical narratives. 
This optimized alignment effectively constitutes the reliable cognitive 
engine that drives the subsequent landslide identification and reasoning 
agents.
\subsection{Collaborative Execution Layer: Designing the Domain-Rule-Enhanced LandslideAgent}\label{subsec:landslideagent-design}
Comprehensive landslide interpretation is an inherently 
composite task, encompassing spatial localization, 
taxonomic classification, morphological extraction, and 
causal inference. Traditionally, these heterogeneous 
sub-tasks are orchestrated either via manual expert 
intervention or through rigid, hard-coded workflows. 
Such static paradigms fundamentally lack the agility 
required to accommodate dynamic tool scheduling and the 
shifting perceptual demands of complex hazard scenarios. 
To transcend these systemic bottlenecks and achieve an 
autonomous analytical paradigm, we propose a 
domain-rule-enhanced agent architecture, termed 
LandslideAgent.

\begin{figure}[htbp]
\centering
\includegraphics[width=\textwidth]{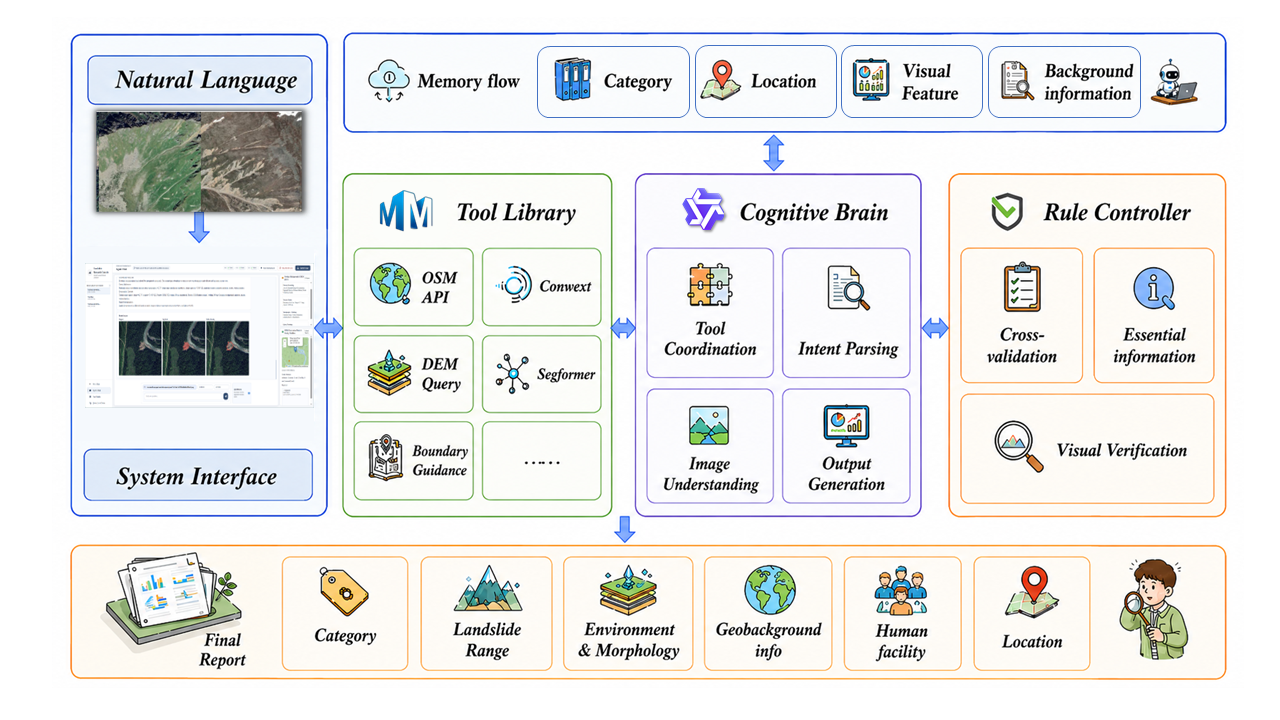}
\caption{Architecture of the proposed LandslideAgent framework}
\label{fig:agent-structure}
\end{figure}

Following the ReAct (Reasoning and Acting) paradigm~\citep{yao2022react}, 
LandslideAgent executes complex tasks through a 
continuous ``Thought-Action-Observation'' cycle. As 
illustrated in Figure~\ref{fig:agent-structure}, the LandslideAgent is driven 
by four synergistic components: a cognitive engine, a 
heterogeneous tool library, a memory management module, 
and a domain rule controller. Serving as the central 
processing unit, the cognitive engine (powered by the 
fine-tuned LandslideVLM detailed in Section~\ref{subsec:landslidevlm-finetuning}) is 
responsible for global scene comprehension, progressive 
task decomposition, optimal tool dispatching, 
and multimodal state synthesis. The tool library 
encapsulates specialized landslide analysis capabilities 
into independent, callable nodes, such as image analysis,
landslide candidate region segmentation, boundary 
refinement, type discrimination, terrain and geological 
querying, surrounding element retrieval, and 
comprehensive assessment. Concurrently, 
the memory management module supersedes conventional 
linear data flows. It dynamically updates, stores, and 
propagates multi-source evidence across consecutive 
reasoning rounds using a structured dictionary format, 
ensuring robust contextual continuity.

A critical departure from conventional, general-purpose 
agents is the integration of the domain rule controller. 
Unconstrained large models frequently exhibit 
unpredictable, open-ended reasoning and redundant 
tool-calling behaviors. To counteract this, the 
controller explicitly translates heuristic 
geoscientific knowledge into deterministic logical 
rules, imposing strict operational boundaries on the 
agent's action space. Specifically tailored to 
guarantee the rigor of the final structured report, 
the LandslideAgent implements a novel dual-domain rule 
constraint mechanism.

The first mechanism enforces an execution constraint 
driven by metadata dependencies. Specifically, 
LandslideAgent formulates the structured fields of the 
final analytical report as a predefined objective 
function with strict structural prerequisites. 
Consequently, to fulfill these parameters, the agent is 
logically compelled to acquire critical intermediate 
representations---such as pixel-level masks, taxonomic 
labels, and geographical contexts---by querying the 
corresponding tool nodes. While the cognitive engine 
LandslideVLM orchestrates the optimal sequence of these 
invocations, the overarching rule ensures that the hard 
dependencies for report generation are rigorously 
satisfied, preventing the agent from skipping essential 
analytical steps.

The second mechanism introduces a recognition 
constraint based on cross-model validation. During 
inference, LandslideAgent requires cross-verification 
between the macroscopic scene perception generated by 
the VLM and the fine-grained extraction outputs of a 
dedicated semantic segmentation model. This stage uses 
a recall-prioritised heuristic. This means that any 
sample supported by either model is retained in the 
candidate pool, thereby minimising the probability of 
missed detections. Furthermore, if the detected hazard 
area falls below a predefined spatial threshold, the 
controller triggers a mandatory boundary 
overlay (superimposing the segmented contours) and 
prompts the VLM to perform a secondary visual 
reappraisal. This ensures that sufficient contextual 
evidence is gathered to support rigorous report synthesis.

\begin{figure}[htbp]
\centering
\includegraphics[width=\textwidth]{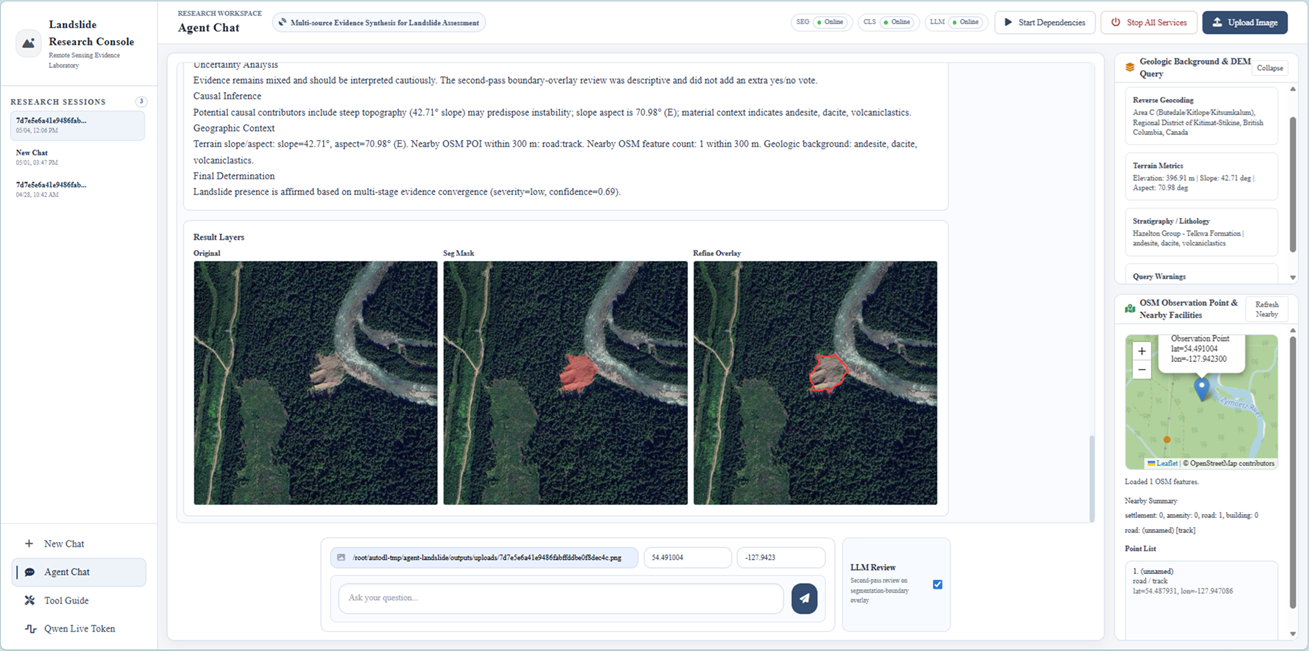}
\caption{System interface of the LandslideAgent platform, showing the task session panel, natural language interaction workspace, reasoning trajectory visualization, and geospatial context rendering viewport}
\label{fig:system-interface}
\end{figure}

At the system implementation level, LandslideAgent 
is constructed upon a highly decoupled, modular 
architecture. The backend infrastructure is anchored 
by the FastAPI asynchronous framework and Uvicorn, 
with the fine-tuned LandslideVLM deployed via PyTorch. 
To facilitate robust geospatial operations, the agent 
is equipped with specialized computational 
libraries (e.g., OpenMMLab for SegFormer and ConvNeXt~\citep{contributors2023openmmlab,contributors2020mmsegmentation}; 
rasterio for spatial data) and interfaces with external 
geographic web services (including OpenTopography, 
Macrostrat, and OpenStreetMap). This design enables the 
autonomous retrieval of multi-source spatial contexts. 
As illustrated in Figure~\ref{fig:system-interface}, the frontend employs the 
Leaflet framework to enable dynamic interaction with 
geospatial data. The user interface is logically 
partitioned into three functional zones: the left panel 
manages historical task sessions and system monitoring 
menus; the central workspace processes natural language 
instructions and remote sensing inputs, visualizing the 
agent's reasoning trajectory and intermediate outputs in 
real time; and the right panel serves as a dedicated 
viewport for rendering the retrieved multi-source 
geographic contexts.

\section{Experiments}\label{sec3}
To systematically evaluate the proposed framework, 
this section designs experiments across three 
dimensions. First, we establish quantitative baselines 
on the LandslideBench dataset using mainstream deep 
learning algorithms for landslide scene classification 
and semantic segmentation to verify its utility as a 
foundational training resource. Second, we assess the 
domain adaptability of LandslideVLM by comparing its 
visual recognition and semantic generation capabilities 
before and after geoscientific fine-tuning. Finally, 
an empirical case study of LandslideAgent demonstrates 
its operational efficacy in heterogeneous tool 
orchestration, logical reasoning, and structured 
report generation for real-world landslide scenarios.
\subsection{Experimental Setup}\label{subsec:experimental-setup}
In the experiments, the LandslideBench dataset is 
strictly divided into training, validation, and test 
sets at an 8:1:1 ratio. For baseline reproducibility, 
all vision models are implemented using the OpenMMLab 
frameworks (MMPreTrain for classification 
and MMSegmentation for segmentation), with input 
images standardized to 512$\times$512 pixels. For the scene 
classification task, models are initialized with 
ImageNet-1K weights and trained for 50 epochs with a 
batch size of 16, and performance is quantified using 
Top-1 and Top-2 Accuracy. Top-k Accuracy measures the 
proportion of test samples for which the ground-truth 
class is included in the model's top-k predicted classes.

For the semantic segmentation task, models are similarly trained for 50 
epochs with a batch size of 8. Segmentation efficacy 
is evaluated via the mean Intersection over Union (mIoU) 
and mean Dice coefficient (mDice)~\citep{sener_landslidesegnet_2024}, defined as 
follows, where $A_{\mathrm{pred}}$ and $A_{\mathrm{gt}}$ denote the predicted 
landslide area and the ground-truth area, respectively.
\begin{equation}
\mathrm{IoU} =
\frac{\left|A_{\mathrm{pred}} \cap A_{\mathrm{gt}}\right|}
{\left|A_{\mathrm{pred}} \cup A_{\mathrm{gt}}\right|}
\end{equation}

\begin{equation}
\mathrm{Dice} =
\frac{2\left|A_{\mathrm{pred}} \cap A_{\mathrm{gt}}\right|}
{\left|A_{\mathrm{pred}}\right| + \left|A_{\mathrm{gt}}\right|}
\end{equation}

To evaluate the semantic generation quality 
of LandslideVLM, we exclusively target four 
open-ended fields: morphological characteristics, 
material composition, movement features, and 
environmental background. Structured 
fields (e.g., landslide type and relative coordinates) 
are excluded from this specific metric, as they are 
deterministically populated by the agent's integrated 
tools. This evaluation is performed strictly on 
landslide-positive samples within the test set. 
We employ a pre-trained sentence 
transformer (all-MiniLM-L6-v2)~\citep{wang2020minilm} to encode both the 
expert-annotated ground truth and the model-generated 
descriptions into dense vector embeddings. The semantic 
consistency per sample is then measured via cosine 
similarity between these vectors, with the final score 
representing the average similarity across the test set. 
To ensure evaluation rigor, false-negative 
detections (i.e., failure to recognize a landslide) 
are strictly penalized by assigning a semantic score 
of 0. Compared to traditional lexical-overlap 
metrics (e.g., BLEU, ROUGE), this embedding-based 
evaluation more accurately captures the semantic 
fidelity of complex geoscientific descriptions.
\subsection{Performance Benchmarking on LandslideBench}\label{subsec:landslidebench-benchmarking}
\subsubsection{Baselines for Fine-grained Landslide Scene Classification}\label{subsubsec:classification-baselines}
To establish baselines for fine-grained landslide scene classification on 
LandslideBench, we evaluated a representative suite 
of visual backbone networks, encompassing both CNN and 
Transformer architectures, such as ResNet-50~\citep{he2016deep}, Swin 
Transformer~\citep{liu2021swin}, CSPDarkNet~\citep{wang2020cspnet}, ConvNeXt~\citep{liu2022convnet}, and EfficientNet~\citep{tan2019efficientnet}. 

\begin{table}[htbp]
\centering
\caption{Performance of Mainstream Classification Baseline Models on the Dataset}
\label{tab:classification-baselines}
\begin{tabular}{lcc}
\hline
\textbf{Model} & \textbf{Top-1 Accuracy} & \textbf{Top-2 Accuracy} \\
\hline
ResNet-50 & 62.21\% & 80.36\% \\
Swin Transformer & 73.97\% & 87.61\% \\
CSPDarkNet & 73.06\% & 86.76\% \\
ConvNeXt & 79.00\% & 95.43\% \\
EfficientNet & 76.71\% & 91.32\% \\
\hline
\end{tabular}
\end{table}

As summarized in Table~\ref{tab:classification-baselines}, while the Top-1 accuracy of 
these models ranges from 62.21\% to 79.00\%, their 
Top-2 accuracy exhibits a pronounced margin of 
improvement. For instance, ConvNeXt increases from 
79.00\% to 95.43\%. This substantial performance 
disparity highlights the severe inter-class visual 
ambiguity inherent in relying exclusively on unimodal 
RGB optical imagery for fine-grained landslide taxonomy. 
Specifically, landslides with distinct kinematic 
mechanisms or material compositions (e.g., debris 
flow vs. earth flow, rock fall vs. rock slide) 
frequently share highly analogous textural patterns and 
morphological contours under 2D optical projection. In 
the absence of 3D topographic priors and macroscopic 
geological context, pure vision models are highly 
susceptible to confusion among these visually 
overlapping categories. This inherent bottleneck of 
unimodal perception objectively underscores the 
necessity of designing LandslideAgent, which mitigates 
these ambiguities through the orchestration of 
multi-source geospatial tools.
\subsubsection{Baselines for Landslide Semantic Segmentation}\label{subsubsec:segmentation-baselines}
Table~\ref{tab:segmentation-baselines} summarizes the semantic segmentation baselines 
evaluated on LandslideBench, encompassing mainstream 
architectures such as DeepLabV3+~\citep{chen2018encoder}, PSPNet~\citep{zhao2017pyramid}, FCN~\citep{long2015fully}, 
SegFormer~\citep{xie2021segformer}, and OCRNet~\citep{yuan2020object}. At a macroscopic level, all 
evaluated models achieve high global metrics, with 
mIoU and mDice generally exceeding 84\% and 90\%, 
respectively. However, this apparent strong performance 
is heavily skewed by class imbalance; the 
straightforward delineation of extensive, homogeneous 
background regions (e.g., intact woodlands, large water 
bodies) artificially inflates the overall scores.

\begin{table}[htbp]
\centering
\caption{Performance of Mainstream Semantic Segmentation Baseline Models on the Dataset}
\label{tab:segmentation-baselines}
\begin{tabular}{lcccc}
\hline
\textbf{Model} & \textbf{mIoU} & \textbf{Landslide-IoU} & \textbf{mDice} & \textbf{Landslide-Dice} \\
\hline
DeepLabV3$+$-ResNet-50 & 84.77\% & 73.80\% & 91.37\% & 84.92\% \\
PSPNet-ResNet-50 & 84.19\% & 72.78\% & 91.00\% & 84.24\% \\
FCN-ResNet-50 & 84.04\% & 72.53\% & 90.90\% & 84.08\% \\
SegFormer-MiT-B2 & 85.39\% & 76.49\% & 92.36\% & 86.68\% \\
OCRNet-HRNet32 & 85.52\% & 75.02\% & 91.85\% & 85.73\% \\
\hline
\end{tabular}
\end{table}

Conversely, when isolating the target landslide class, performance is significantly degraded. 
The Landslide-IoU across models exhibits a decline, 
ranging from 72.53\% to 76.49\%, accompanied by a 
decrease in the corresponding Landslide-Dice, from 
84.08\% to 86.68\%. This degradation underscores the 
inherent challenge of landslide boundary delineation. 
The irregular morphology of landslides, in conjunction 
with subtle visual transitions at their margins, where 
slide areas frequently blend into surrounding bare soil, 
weathered rock, or disturbed vegetation, severely 
hinders precise pixel-wise classification.

Architecturally, while OCRNet holds a marginal edge in global mIoU, 
SegFormer demonstrates superior robustness for the 
specific task of landslide extraction, achieving the 
peak Landslide-IoU (76.49\%) and Landslide-Dice (86.68\%). 
This superiority is primarily attributed to SegFormer's 
Transformer-based self-attention mechanism, which 
effectively models long-range spatial dependencies. 
By capturing broader macro-contextual cues, it better 
constrains the highly ambiguous landslide boundaries 
against complex heterogeneous backgrounds. Nevertheless, 
despite optimal utilization of visual context, 
SegFormer's performance ceiling (Landslide-IoU $<$ 77\%) 
indicates that relying solely on unimodal optical 
features remains insufficient for precise landslide 
delimitation. This further corroborates the premise 
that explicit geometric constraints and multi-source 
geological knowledge, as operationalized by 
LandslideAgent, are indispensable for breaking 
through current segmentation bottlenecks.

Overall, these benchmark evaluations demonstrate that 
LandslideBench possesses high model discriminability, 
effectively capturing the performance disparities among 
various feature-modeling paradigms when confronting 
geospatially complex characteristics. While this 
establishes a robust evaluative data benchmark for 
intelligent landslide interpretation, the universally 
observed performance ceilings across all pure-vision 
baselines deliver a clear message: solely relying on 
unimodal optical representation is fundamentally 
insufficient. It is exactly this inherent limitation 
revealed by LandslideBench that motivates our 
transition from traditional end-to-end networks to 
the proposed LandslideAgent, paving the way for 
multi-source knowledge orchestration to break through 
current modality barriers.

\subsection{Performance Evaluation on LandslideVLM}\label{subsec:landslidevlm-performance}
\subsubsection{VLM Baselines for Hierarchical Landslide Classification}\label{subsubsec:vlm-classification}

As shown in Table~\ref{tab:vlm-classification}, LandslideVLM outperforms its 
zero-shot foundational counterpart, 
Qwen3-VL-8B-Instruct, in several key areas. 
Specifically, in the binary landslide scene 
classification task, LandslideVLM's accuracy 
on the LandslideBench test set increases from 85.84\% to 96.80\%. 
Even more notably, in the structurally complex 
fine-grained classification task, accuracy increases 
from 19.18\% to 52.05\%, representing an 
impressive absolute gain of 32.87 percentage points. 
The empirical results demonstrate that fine-tuning 
generalized VLMs on LandslideBench via the LoRA 
paradigm effectively calibrates their visual 
representation mechanisms, enabling the robust 
extraction of domain-specific geospatial features 
inherent to landslides. 

\begin{table}[htbp]
\centering
\caption{Landslide Classification Performance Before and After Fine-Tuning}
\label{tab:vlm-classification}
\begin{tabular}{lcc}
\hline
\textbf{Task} & \textbf{Qwen3-VL-8B-Instruct} & \textbf{LandslideVLM} \\
\hline
Binary landslide scene classification & 85.84\% & 96.80\% \\
Fine-grained landslide scene classification & 19.18\% & 52.05\% \\
\hline
\end{tabular}
\end{table}

However, an objective cross-comparison reveals a 
critical bottleneck. Despite the substantial relative 
gains achieved through fine-tuning, LandslideVLM's 
absolute accuracy in fine-grained classification (52.05\%) 
still lags significantly behind the dedicated 
pure-vision models evaluated in Section~\ref{subsec:landslidebench-benchmarking}. This 
performance gap highlights an inherent limitation 
of current multimodal large architectures: a deficiency 
in perceptual granularity when confronting tasks heavily 
reliant on localized, high-frequency textural features. 
It demonstrates that while VLMs excel at macroscopic 
semantic comprehension, they cannot yet entirely 
supplant specialized interpretative models for 
microscopic detail extraction.

\subsubsection{Baselines for Landslide Semantic Description}\label{subsubsec:vlm-description}

Quantitative evaluations of semantic description 
quality (Table~\ref{tab:vlm-description}) indicate that LandslideVLM 
consistently surpasses its zero-shot baseline, 
Qwen3-VL-8B-Instruct, across all four predefined 
descriptive dimensions. Specifically, LandslideVLM 
yields improved scores of 0.4910, 0.6484, 0.4605, and 
0.7082 in morphological features, material composition, 
movement characteristics, and environmental context, 
respectively, thereby elevating the overall average 
score from 0.4179 to 0.5770. These metrics reveal 
that domain-specific fine-tuning significantly enhances 
the model's capacity to extract multi-level geological 
semantics. The pronounced improvements in the material 
composition and environmental context dimensions suggest 
that LandslideVLM has effectively learned the complex 
mapping mechanism between static visual textures in 
remote sensing imagery and their underlying geospatial 
semantics.

\begin{table}[htbp]
\centering
\caption{Landslide Semantic Description Performance Before and After Fine-Tuning}
\label{tab:vlm-description}
\begin{tabular}{lcc}
\hline
\textbf{Metric} & \textbf{Qwen3-VL-8B-Instruct} & \textbf{LandslideVLM} \\
\hline
Morphological Features & 0.3423 & 0.4910 \\
Material Composition & 0.4506 & 0.6484 \\
Movement Characteristics & 0.3841 & 0.4605 \\
Environmental Context & 0.4947 & 0.7082 \\
Average & 0.4179 & 0.5770 \\
\hline
\end{tabular}
\end{table}

However, it is noteworthy that the performance gain in 
describing movement characteristics remains 
comparatively marginal (an increase of approximately 
0.08). This limitation strictly aligns with geological 
realities: landslide kinematics and deformation 
mechanisms are inherently dynamic processes governed 
by temporal evolution. Consequently, inferring complex 
kinematic behaviors solely from single-epoch static 
imagery presents an inherent physical constraint, 
rendering it an ill-posed challenge even for advanced 
multimodal architectures.

Synthesizing the evaluations of both hierarchical classification and 
semantic description, it becomes evident that while 
fine-tuning effectively empowers the VLM with 
domain-specific knowledge, relying solely on a 
single unified architecture remains insufficient 
for comprehensive landslide interpretation. As 
demonstrated above, the VLM exhibits robust 
capabilities in macro-level semantic mapping and 
geographical context reasoning, yet it encounters 
inherent bottlenecks when confronted with tasks 
requiring localized, fine-grained visual perception. 
This capability dichotomy directly motivates the design 
philosophy of the proposed LandslideAgent. Rather than 
forcing the VLM to perform complex pixel-level feature 
extraction, the Agent framework decouples the 
interpretation pipeline. It leverages the VLM's 
validated semantic strengths as a central logical 
planner for high-level reasoning and task formulation, 
while dynamically integrating specialized vision 
models (as evaluated in Section~\ref{subsec:landslidebench-benchmarking}) as external tools 
to handle high-precision visual perception. This 
collaborative paradigm effectively bypasses the 
inherent perceptual limitations of large multimodal 
models, fully capitalizing on their cognitive reasoning 
advantages to ensure the accuracy and robustness of 
the ultimate interpretation framework.

\subsection{LandslideAgent Case Study}
To evaluate the operational capability of the 
proposed LandslideAgent, a representative landslide 
event in British Columbia, Canada, was selected for a 
case study. This scenario demonstrates the system's 
proficiency in orchestrating collaborative perception, 
dynamic tool invocation, spatial reasoning, and 
multi-source knowledge integration within a realistic 
hazard interpretation workflow.

\begin{figure}[htbp]
\centering
\includegraphics[width=\textwidth]{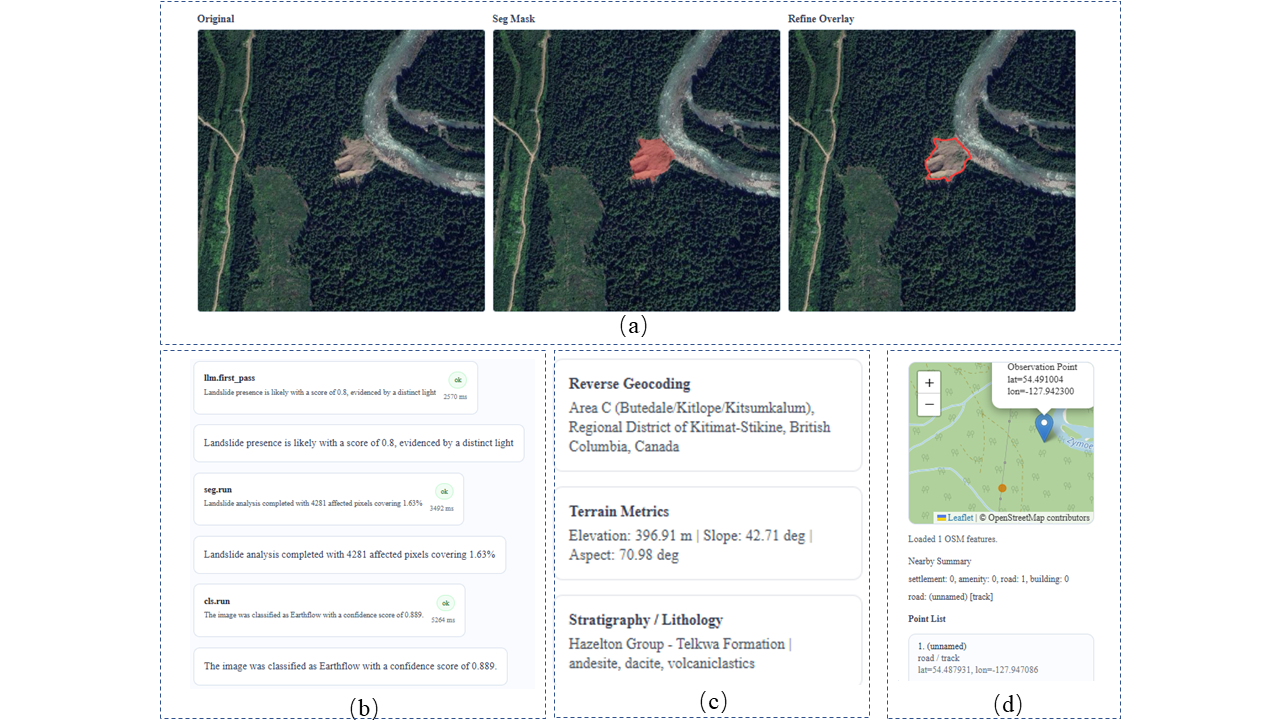}
\caption{Visual boundary highlighting and interpretation results generated during the LandslideAgent case study}
\label{fig:case-boundary}
\end{figure}

Upon task initiation, LandslideAgent orchestrates 
LandslideVLM and the dedicated semantic segmentation 
model (Section~\ref{subsec:landslidebench-benchmarking}) to conduct preliminary 
detection (Figure~\ref{fig:case-boundary}a), with a cross-validation 
mechanism confirming the initial presence of a 
landslide. Crucially, to ensure analytical accuracy, 
the LandslideAgent does not blindly execute a rigid 
pipeline. Instead, it evaluates the specific situational 
context to dynamically trigger appropriate verification 
rules. By quantitatively analyzing the segmentation 
mask, the LandslideAgent identifies that the hazard 
footprint occupies merely 1.63\% of the overall scene. 
Recognizing this as a high-risk scenario where small 
targets are easily obscured by background noise, the 
LandslideAgent automatically triggers a visual 
calibration workflow. It translates the spatial 
mask into bounding visual prompts and feeds them 
back into the reasoning chain. This condition-triggered 
verification mechanism explicitly enforces precise 
visual grounding, guaranteeing that subsequent 
fine-grained semantic extractions remain strictly 
focused on the target area and unaffected by global 
interference.

Following spatial localization, the LandslideAgent transitions into a multi-dimensional 
feature extraction phase via autonomous tool invocation. 
It first calls the fine-grained classification tool, 
categorizing the landslide as an earth flow (Figure~\ref{fig:case-boundary}b). 
Subsequently, to contextualize the event within its 
physical geography, the LandslideAgent triggers the 
geospatial background retrieval tool, automatically 
parsing regional DEM and geological datasets. The 
extracted geomorphometric parameters reveal a slope 
of 42.71$^\circ$ and an aspect of 70.98$^\circ$ (Figure~\ref{fig:case-boundary}c), which 
are structurally consistent with the illumination and 
shading patterns observed in the optical imagery.

For exposure and risk assessment, the LandslideAgent 
invokes the OSM query and spatial topology analysis 
tools to identify nearby critical infrastructure. 
The system successfully flags a vulnerable road segment 
situated within a 300-meter buffer zone of the 
delineated landslide boundary, automatically generating 
a localized risk alert (Figure~\ref{fig:case-boundary}d).

Having aggregated these heterogeneous multi-source data, 
the LandslideAgent orchestrates a second-pass visual 
comprehension utilizing the generated spatial prompts. 
Constrained by the explicitly bounded landslide 
footprint, LandslideVLM successfully suppresses global 
background interference. It accurately delineates the 
macro-spatial context---situating the failure adjacent to 
a river channel in the upper-middle sector of the 
scene---while simultaneously generating a micro-level 
semantic description. The model characterizes the 
landslide body as predominantly light brown, featuring 
smooth surface morphology and indistinct boundaries 
without severe runoff erosion signatures. This visually 
derived semantic profile directly corroborates 
the tool-inferred earth flow classification, 
establishing a coherent observational basis for 
kinematic analysis.

Ultimately, the LandslideAgent synthesizes the modular 
outputs to automatically compile a structured, 
multi-section geohazard assessment report (Appendix~\ref{secA1}). 
This report provides analytical conclusions, evidence 
summaries, spatial-typological interpretations, exposure 
contexts, and an explicit uncertainty evaluation. 
This operational case underscores LandslideAgent's 
capacity to bridge coarse-grained localization, 
multi-source parameter extraction, and cross-modal 
reasoning. By successfully synergizing the high-level 
semantic reasoning of VLMs with the pixel-level precision 
of specialized external tools, the proposed architecture 
effectively bridges the gap between broad cognitive 
understanding and rigorous visual perception 
in geohazard analysis.

\section{Conclusion}
This study introduces an instruction-driven agentic 
framework to resolve the longstanding disconnect between 
visual perception and high-level geoscientific reasoning 
in traditional remote sensing-based landslide 
interpretation. As a foundational step, we 
curated LandslideBench, a fine-grained multimodal 
dataset coupling high-resolution imagery with 
pixel-level mask and multi-dimensional textual 
attributes, enabling systematic benchmark evaluations 
of contemporary visual models. To suppress 
domain-specific hallucinations inherent in 
general-purpose vision-language models, 
we developed LandslideVLM via parameter-efficient 
LoRA fine-tuning. Evaluations confirm that this 
specialized model yields substantial improvements 
in binary and fine-grained landslide scene 
classification, alongside a marked enhancement 
in generating expert-level descriptions of material 
composition and environmental contexts. Building 
upon these advancements, we proposed LandslideAgent 
augmented by dual domain rules. Empirical findings 
demonstrate the agent's proficiency in autonomous 
cross-validation and spatial reasoning across 
multi-source data, circumventing the rigidities 
of single-model pipelines to execute a fully 
automated landslide analysis.

Despite these advancements in automating hazard 
interpretation, our findings delineate the inherent 
cognitive boundaries of the current static reasoning 
paradigm. On the one hand, pervasive spectral 
confusion---driven by high intra-class variance and 
inter-class similarity---imposes a persistent 
bottleneck on fine-grained classification, 
restricting the agent's perceptual limits on 
single-temporal imagery. On the other hand, the 
marginal improvements observed in inferring 
landslide kinematic characteristics underscore 
that relying on single-frame static snapshots 
fundamentally restricts the agent's capacity to 
reconstruct complex hazard dynamics. Our future work 
will transition LandslideAgent from a static scene 
interpreter to a dynamic, continuous-monitoring 
intelligent system. Rather than merely expanding 
the data dimension, we intend to augment the agent's 
tool library with multi-source time-series 
observations (e.g., InSAR-derived surface deformation) 
and process-based geological models. Consequently, 
advancing this agent-driven paradigm provides a scalable 
mechanism to synthesize spatiotemporal data and complex 
geoscientific rules, facilitating continuous and 
quantitative geohazard analysis.

\section*{Statements and Declarations}

\begin{itemize}
\item \textbf{Conflict of interest/Competing interests:}
The authors declare that they have no competing interests.

\item \textbf{Ethics approval and consent to participate:}
Not applicable.

\item \textbf{Consent for publication:}
Not applicable.

\item \textbf{Data availability:}
The data that support the findings of this study will be made available at \url{https://github.com/GeoRSAI/LandslideAgent}.

\item \textbf{Materials availability:}
Not applicable.

\item \textbf{Code availability:}
The code used in this study will be made available at \url{https://github.com/GeoRSAI/LandslideAgent}.

\item \textbf{Author contribution:}
Chengfu Liu contributed to conceptualization, methodology, experiments, data analysis, and manuscript drafting. Dongyang Hou contributed to supervision, conceptualization, project administration, and manuscript revision. Junwu Xiang, Cheng Yang, Xuezhi Cui, Zeyuan Wang, Liangtian Liu, and Zelang Miao contributed to data preparation, experimental validation, result analysis, and manuscript revision. All authors reviewed and approved the final manuscript.
\end{itemize}

\begin{appendices}

\refstepcounter{section}
\section*{Appendix~\thesection. Instance of agent-generated report}\label{secA1}
\addcontentsline{toc}{section}{Appendix~\thesection. Instance of agent-generated report}

\begingroup

\small

\subsubsection*{Conclusion}

Landslide presence is confirmed by cross-stage evidence. A small earth flow landslide is confirmed with low severity, characterized by a discrete mass of light brown soil and fragmented rock displaced from a forested hillside into a meandering river channel, causing localized disruption to vegetative continuity.

\subsubsection*{Evidence Summary}

Initial Screening: Confirmed landslide presence, citing A distinct landslide mass with light brown exposed soil and fragmented rock is visible near a meandering river channel; the slide appears to have originated from a forested hillside adjacent to the watercourse, showing clear displacement of terrain materials into the river's path, disrupting natural vegetation continuity along its edges. Segmentation Results: 1 landslide polygon(s) identified, covering 1.6\% of image area (4281 pixels). Region Refinement: 1 retained candidate region(s), summarized as regions=1. Classification Reference: Tentatively labeled ``earth flow'' (confidence: 89\%). This label is used only for contextual support. Consistency Check: Multi-stage outputs are spatially aligned, and the second-pass boundary-overlay review was used as descriptive support only. Second-pass Review: This workflow performed a second-pass VLM review on the full image with segmentation boundaries overlaid, explicitly supplementing the spatial description of 1 highlighted region(s), after first-pass screening and segmentation-guided refinement already indicated landslide presence. Review note: The red-outlined landslide area is located towards the center-right of the image frame, adjacent to the meandering river channel on its right side, with the river flowing through dense forested terrain while the landslide mass exhibits disrupted surface texture indicating material movement.

\subsubsection*{Image and Spatial Interpretation}

Overall scene description: A distinct landslide mass with light brown exposed soil and fragmented rock is visible near a meandering river channel; the slide appears to have originated from a forested hillside adjacent to the watercourse, showing clear displacement of terrain materials into the river's path, disrupting natural vegetation continuity along its edges. Localized linear distribution concentrated around the interface between the forested hillside and the river channel, extending slightly downstream into the water body. Primary candidate region lies in the middle-center part of the frame; bbox=[244.0, 197.0, 326.0, 275.0] pixels. Segmentation footprint covers 1.6\% of the frame, with candidate-area ratio 0.0163. Detailed image interpretation: The image shows a distinct landslide mass with light brown exposed soil and angular rock fragments situated near a winding river channel; the slide originates from a densely forested slope adjacent to the watercourse, where displaced materials extend into the river's path, interrupting the natural edge of the riparian zone and creating a visibly disturbed terrain pattern against the surrounding intact vegetation.

\subsubsection*{Landslide Typology (Reference Only)}

Reference classification: earth flow (0.89). Classification adheres strictly to the model output assigning `earth flow' with top-tier confidence (0.889), prioritizing observed morphology over competing categories such as debris flow or rock fall.

\subsubsection*{Geographic and Exposure Context}

Potential environmental and infrastructure exposure exists around 1 mapped nearby feature(s) within 300 m. Potential causal contributors include steep topography (42.71$^\circ$ slope) may predispose instability; slope aspect is 70.98$^\circ$ (E); material context indicates andesite, dacite, volcaniclastics. Terrain slope/aspect: slope=42.71$^\circ$, aspect=70.98$^\circ$ (E). Nearby OSM POI within 300 m: road:track. Nearby OSM feature count: 1 within 300 m. Geologic background: andesite, dacite, volcaniclastics.

\subsubsection*{Reliability and Uncertainty}

Decision reliability is described qualitatively from cross-stage agreement; no calibrated overall confidence score is reported. The second-pass boundary-overlay review was descriptive and did not add an extra yes/no vote. Uncertainty arises from limited spatial extent of the landslide relative to the full image context, which may affect precise boundary delineation despite successful isolation via segmentation; however, the presence of consistent morphological features supports reliable identification.

\subsubsection*{Final Determination}

Landslide presence is affirmed based on multi-stage evidence convergence (severity=low).

\endgroup

\end{appendices}


\bibliography{sn-bibliography}% common bib file

@article{alcantara-ayala_landslides_2025,
	title = {Landslides in a changing world},
	volume = {22},
	issn = {1612-5118},
	url = {https://doi.org/10.1007/s10346-024-02451-1},
	doi = {10.1007/s10346-024-02451-1},
	number = {9},
	journal = {Landslides},
	author = {Alc{\'a}ntara-Ayala, Irasema},
	month = sep,
	year = {2025},
	pages = {2851--2865},
}

@article{areerob2025multimodal,
  author = {Areerob, Kittitouch and Nguyen, Van-Quang and Li, Xianfeng and Inadomi, Shogo and Shimada, Toru and Kanasaki, Hiroyuki and Wang, Zhijie and Suganuma, Masanori and Nagatani, Keiji and Chun, Pang-jo and Okatani, Takayuki},
  title = {Multimodal artificial intelligence approaches using large language models for expert-level landslide image analysis},
  journal = {Computer-Aided Civil and Infrastructure Engineering},
  volume = {40},
  number = {19},
  pages = {2900--2921},
  year = {2025},
  doi = {10.1111/mice.13482}
}

@misc{bai2025qwen3,
  author = {Bai, Shuai and Cai, Yuxuan and Chen, Ruizhe and Chen, Keqin and Chen, Xionghui and Cheng, Zesen and Deng, Lianghao and Ding, Wei and Gao, Chang and Ge, Chunjiang and Ge, Wenbin and Guo, Zhifang and Huang, Qidong and Huang, Jie and Huang, Fei and Hui, Binyuan and Jiang, Shutong and Li, Zhaohai and Li, Mingsheng and Li, Mei and Li, Kaixin and Lin, Zicheng and Lin, Junyang and Liu, Xuejing and Liu, Jiawei and Liu, Chenglong and Liu, Yang and Liu, Dayiheng and Liu, Shixuan and Lu, Dunjie and Luo, Ruilin and Lv, Chenxu and Men, Rui and Meng, Lingchen and Ren, Xuancheng and Ren, Xingzhang and Song, Sibo and Sun, Yuchong and Tang, Jun and Tu, Jianhong and Wan, Jianqiang and Wang, Peng and Wang, Pengfei and Wang, Qiuyue and Wang, Yuxuan and Xie, Tianbao and Xu, Yiheng and Xu, Haiyang and Xu, Jin and Yang, Zhibo and Yang, Mingkun and Yang, Jianxin and Yang, An and Yu, Bowen and Zhang, Fei and Zhang, Hang and Zhang, Xi and Zheng, Bo and Zhong, Humen and Zhou, Jingren and Zhou, Fan and Zhou, Jing and Zhu, Yuanzhi and Zhu, Ke},
  title = {{Qwen3-VL} Technical Report},
  year = {2025},
  eprint = {2511.21631},
  archivePrefix = {arXiv},
  primaryClass = {cs.CV}
}

@article{casagli2023landslide,
  author = {Casagli, Nicola and Intrieri, Emanuele and Tofani, Veronica and Gigli, Giovanni and Raspini, Federico},
  title = {Landslide detection, monitoring and prediction with remote-sensing techniques},
  journal = {Nature Reviews Earth \& Environment},
  volume = {4},
  number = {1},
  pages = {51--64},
  year = {2023},
  doi = {10.1038/s43017-022-00373-x}
}

@inproceedings{chen2018encoder,
  author = {Chen, Liang-Chieh and Zhu, Yukun and Papandreou, George and Schroff, Florian and Adam, Hartwig},
  title = {Encoder-Decoder with Atrous Separable Convolution for Semantic Image Segmentation},
  booktitle = {Proceedings of the European Conference on Computer Vision (ECCV)},
  year = {2018},
  pages = {801--818}
}

@article{chen2026integration,
  author = {Chen, Zhaohui and Asadi Shamsabadi, Elyas and Jiang, Sheng and Shen, Luming and Dias-da-Costa, Daniel},
  title = {Integration of large vision language models for efficient post-disaster damage assessment and reporting},
  journal = {Nature Communications},
  volume = {17},
  number = {1},
  pages = {1481},
  year = {2026},
  doi = {10.1038/s41467-025-68216-z}
}

@article{fang_rapid_2026,
	title = {Rapid and robust landslide mapping from optical {EO} imagery using a mamba-based deep learning framework},
	issn = {1612-5118},
	url = {https://doi.org/10.1007/s10346-026-02789-8},
	doi = {10.1007/s10346-026-02789-8},
	journal = {Landslides},
	author = {Fang, Chengyong and Fan, Xuanmei and Wang, Xin and Bhuyan, Kushanav and Dou, Xiangyang and Zhong, Hao and Xia, Mingyao and Catani, Filippo},
	month = may,
	year = {2026},
}

@article{feng_convolutional_2025,
	title = {Convolutional neural network-based deep learning for landslide susceptibility mapping in the {Bakhtegan} watershed},
	volume = {15},
	issn = {2045-2322},
	url = {https://doi.org/10.1038/s41598-025-96748-3},
	doi = {10.1038/s41598-025-96748-3},
	number = {1},
	journal = {Scientific Reports},
	author = {Feng, Li and Zhang, Maosheng and Mao, Yimin and Liu, Hao and Yang, Chuanbo and Dong, Ying and Nanehkaran, Yaser A.},
	month = apr,
	year = {2025},
	pages = {13250},
}

@misc{feng2025earth,
  author = {Feng, Peilin and Lv, Zhutao and Ye, Junyan and Wang, Xiaolei and Huo, Xinjie and Yu, Jinhua and Xu, Wanghan and Zhang, Wenlong and Bai, Lei and He, Conghui and Li, Weijia},
  title = {Earth-Agent: Unlocking the Full Landscape of Earth Observation with Agents},
  year = {2026},
  eprint = {2509.23141},
  archivePrefix = {arXiv},
  primaryClass = {cs.CV}
}

@article{fu2026cnn,
  author = {Fu, Zijin and Wang, Fawu and Zhong, Junfei and Catani, Filippo and Dou, Jie and You, Qi and Zhang, Bo},
  title = {A {CNN}--{Transformer} hybrid network for efficient cross-region landslide detection by transfer learning},
  journal = {Landslides},
  year = {2026},
  doi = {10.1007/s10346-026-02733-w}
}

@article{yunjian20262025,
  author = {Gao, Yunjian and Tie, Yongbo and Li, Zongliang and Ba, Renji and Yin, Chuanjie and Ge, Hua and Li, Pengyue},
  title = {2025 {Dingqing} catastrophic landslide induced by frost heaving on high hillslopes in the southeast {Tibetan} {Plateau}, {China}},
  journal = {Landslides},
  volume = {23},
  number = {6},
  pages = {1625--1648},
  year = {2026},
  doi = {10.1007/s10346-026-02739-4}
}

@article{ghorbanzadeh2022landslide4sense,
  author = {Ghorbanzadeh, Omid and Xu, Yonghao and Ghamisi, Pedram and Kopp, Michael and Kreil, David},
  title = {{Landslide4Sense}: Reference Benchmark Data and Deep Learning Models for Landslide Detection},
  journal = {IEEE Transactions on Geoscience and Remote Sensing},
  volume = {60},
  pages = {1--17},
  year = {2022},
  doi = {10.1109/TGRS.2022.3215209}
}

@inproceedings{he2016deep,
  author = {He, Kaiming and Zhang, Xiangyu and Ren, Shaoqing and Sun, Jian},
  title = {Deep Residual Learning for Image Recognition},
  booktitle = {Proceedings of the IEEE Conference on Computer Vision and Pattern Recognition (CVPR)},
  year = {2016},
  pages = {770--778}
}

@misc{hu2022lora,
  author = {Hu, Edward J. and Shen, Yelong and Wallis, Phillip and Allen-Zhu, Zeyuan and Li, Yuanzhi and Wang, Shean and Wang, Lu and Chen, Weizhu},
  title = {{LoRA}: Low-Rank Adaptation of Large Language Models},
  year = {2021},
  eprint = {2106.09685},
  archivePrefix = {arXiv},
  primaryClass = {cs.CL}
}

@article{hungr2014varnes,
  author = {Hungr, Oldrich and Leroueil, Serge and Picarelli, Luciano},
  title = {The {Varnes} classification of landslide types, an update},
  journal = {Landslides},
  volume = {11},
  number = {2},
  pages = {167--194},
  year = {2014},
  doi = {10.1007/s10346-013-0436-y}
}

@article{ji2020landslide,
  author = {Ji, Shunping and Yu, Dawen and Shen, Chaoyong and Li, Weile and Xu, Qiang},
  title = {Landslide detection from an open satellite imagery and digital elevation model dataset using attention boosted convolutional neural networks},
  journal = {Landslides},
  volume = {17},
  number = {6},
  pages = {1337--1352},
  year = {2020},
  doi = {10.1007/s10346-020-01353-2}
}

@article{liu2025lmhld,
  author = {Liu, Guanting and Wang, Yi and Chen, Xi and Du, Baoyu and Li, Penglei and Wu, Yuan and Fang, Zhice and Ma, Peifeng},
  title = {{LMHLD}: A Large-Scale Multisource High-Resolution Landslide Dataset for Landslide Detection Based on Deep Learning},
  journal = {IEEE Transactions on Geoscience and Remote Sensing},
  volume = {63},
  pages = {1--15},
  year = {2025},
  doi = {10.1109/TGRS.2025.3619062}
}

@inproceedings{liu2021swin,
  author = {Liu, Ze and Lin, Yutong and Cao, Yue and Hu, Han and Wei, Yixuan and Zhang, Zheng and Lin, Stephen and Guo, Baining},
  title = {{Swin Transformer}: Hierarchical Vision Transformer Using Shifted Windows},
  booktitle = {Proceedings of the IEEE/CVF International Conference on Computer Vision (ICCV)},
  pages = {10012--10022},
  year = {2021}
}

@inproceedings{liu2022convnet,
  author = {Liu, Zhuang and Mao, Hanzi and Wu, Chao-Yuan and Feichtenhofer, Christoph and Darrell, Trevor and Xie, Saining},
  title = {A {ConvNet} for the 2020s},
  booktitle = {Proceedings of the IEEE/CVF Conference on Computer Vision and Pattern Recognition (CVPR)},
  pages = {11976--11986},
  year = {2022}
}

@inproceedings{long2015fully,
  author = {Long, Jonathan and Shelhamer, Evan and Darrell, Trevor},
  title = {Fully Convolutional Networks for Semantic Segmentation},
  booktitle = {Proceedings of the IEEE Conference on Computer Vision and Pattern Recognition (CVPR)},
  year = {2015},
  pages = {3431--3440}
}

@article{mancino2025unified,
  author = {Mancino, Saverio and Sblano, Anna and Lovergine, Francesco Paolo and Massimi, Vincenzo and Sethi, Tushar and Capolongo, Domenico and Amatulli, Giuseppe},
  title = {Unified {Global} {Landslide} Catalogue ({UGLC}): A single, standardised global-scale landslide dataset},
  journal = {Earth System Science Data Discussions},
  volume = {2025},
  pages = {1--41},
  year = {2025},
  doi = {10.5194/essd-2025-174}
}

@article{meena2023hr,
  author = {Meena, S. R. and Nava, L. and Bhuyan, K. and Puliero, S. and Soares, L. P. and Dias, H. C. and Floris, M. and Catani, F.},
  title = {{HR}-{GLDD}: a globally distributed dataset using generalized deep learning ({DL}) for rapid landslide mapping on high-resolution ({HR}) satellite imagery},
  journal = {Earth System Science Data},
  volume = {15},
  number = {7},
  pages = {3283--3298},
  year = {2023},
  doi = {10.5194/essd-15-3283-2023}
}

@misc{contributors2023openmmlab,
  author = {{MMPreTrain Contributors}},
  title = {{OpenMMLab}'s Pre-training Toolbox and Benchmark},
  year = {2023},
  howpublished = {\url{https://github.com/open-mmlab/mmpretrain}}
}

@misc{contributors2020mmsegmentation,
  author = {{MMSegmentation Contributors}},
  title = {{MMSegmentation}: {OpenMMLab} Semantic Segmentation Toolbox and Benchmark},
  year = {2020},
  howpublished = {\url{https://github.com/open-mmlab/mmsegmentation}}
}

@inproceedings{ravi2024sam,
  author = {Ravi, Nikhila and Gabeur, Valentin and Hu, Yuan-Ting and Hu, Ronghang and Ryali, Chaitanya and Ma, Tengyu and Khedr, Haitham and R{\"a}dle, Roman and Rolland, Chloe and Gustafson, Laura and Mintun, Eric and Pan, Junting and Alwala, Kalyan Vasudev and Carion, Nicolas and Wu, Chao-Yuan and Girshick, Ross and Doll{\'a}r, Piotr and Feichtenhofer, Christoph},
  title = {{SAM} 2: Segment Anything in Images and Videos},
  booktitle = {International Conference on Learning Representations},
  volume = {2025},
  pages = {28085--28128},
  year = {2025}
}

@article{reghunath2026disasterreliefgpt,
  author = {Reghunath, Lekshmi Chandrika and Abhishek, Athikkal Sudhir and Changat, Arjun and Unnikrishnan, Arjun and Rai, Ayush Kumar and Napoli, Christian and Randieri, Cristian},
  title = {{DisasterReliefGPT}: Multimodal {AI} for Autonomous Disaster Impact Assessment and Crisis Communication},
  journal = {Technologies},
  volume = {14},
  number = {3},
  pages = {179},
  year = {2026},
  doi = {10.3390/technologies14030179}
}

@article{sener_landslidesegnet_2024,
	title = {{LandslideSegNet}: an effective deep learning network for landslide segmentation using remote sensing imagery},
	volume = {17},
	issn = {1865-0481},
	url = {https://doi.org/10.1007/s12145-024-01434-z},
	doi = {10.1007/s12145-024-01434-z},
	number = {5},
	journal = {Earth Science Informatics},
	author = {{\c{S}}ener, Abdullah and Ergen, Burhan},
	month = oct,
	year = {2024},
	pages = {3963--3977},
}

@inproceedings{song2020mpnet,
  author = {Song, Kaitao and Tan, Xu and Qin, Tao and Lu, Jianfeng and Liu, Tie-Yan},
  title = {{MPNet}: Masked and Permuted Pre-training for Language Understanding},
  booktitle = {Advances in Neural Information Processing Systems},
  volume = {33},
  pages = {16857--16867},
  year = {2020}
}

@article{song2025landslide,
  author = {Song, Yuyang and Hao, Lina and Li, Weile},
  title = {Landslide detection using deep learning on remotely sensed images},
  journal = {Applied Computing and Geosciences},
  volume = {27},
  pages = {100278},
  year = {2025},
  doi = {10.1016/j.acags.2025.100278}
}

@InProceedings{Suris_2023_ICCV,
    author    = {Sur{\'\i}s, D{\'\i}dac and Menon, Sachit and Vondrick, Carl},
    title     = {ViperGPT: Visual Inference via Python Execution for Reasoning},
    booktitle = {Proceedings of the IEEE/CVF International Conference on Computer Vision (ICCV)},
    month     = {October},
    year      = {2023},
    pages     = {11888-11898}
}

@misc{talemi2026agenticairemotesensing,
      title={Agentic AI in Remote Sensing: Foundations, Taxonomy, and Emerging Systems}, 
      author={Talemi, Niloufar Alipour and Boone, Julia and Afghah, Fatemeh},
      year={2026},
      eprint={2601.01891},
      archivePrefix={arXiv},
      primaryClass={cs.CV},
}

@inproceedings{tan2019efficientnet,
  author = {Tan, Mingxing and Le, Quoc},
  title = {{EfficientNet}: Rethinking Model Scaling for Convolutional Neural Networks},
  booktitle = {Proceedings of the 36th International Conference on Machine Learning},
  series = {Proceedings of Machine Learning Research},
  volume = {97},
  pages = {6105--6114},
  publisher = {PMLR},
  year = {2019},
  url = {https://proceedings.mlr.press/v97/tan19a.html}
}

@misc{tang2025intelligent,
  title={Intelligent Remote Sensing Agents: A Survey},
  author={Tang, Jiaqi and Yan, Yingying and Wang, Qianzhou and Xia, Yuyang and Geng, Botong and Chen, Jianmin and Ma, Ke and Zhai, Youyang and He, Qingfeng and Shao, Weigeng and Sun, Yunjin and Dai, Junwei and Chen, Chuxi and Xu, Xiaogang and Yao, Kelu and Zhang, Lei and Wei, Wei and Chen, Qifeng and Plaza, Antonio and Zhang, Yanning},
  year={2026},
  howpublished={GitHub repository},
  note={\url{https://github.com/PolyX-Research/Awesome-Remote-Sensing-Agents}}
}

@misc{hong2025glm,
  author = {{GLM-V Team} and Hong, Wenyi and Yu, Wenmeng and Gu, Xiaotao and Wang, Guo and Gan, Guobing and Tang, Haomiao and Cheng, Jiale and Qi, Ji and Ji, Junhui and Pan, Lihang and Duan, Shuaiqi and Wang, Weihan and Wang, Yan and Cheng, Yean and He, Zehai and Su, Zhe and Yang, Zhen and Pan, Ziyang and Zeng, Aohan and Wang, Baoxu and Chen, Bin and Shi, Boyan and Pang, Changyu and Zhang, Chenhui and Yin, Da and Yang, Fan and Chen, Guoqing and Li, Haochen and Zhu, Jiale and Chen, Jiali and Xu, Jiaxing and Xu, Jiazheng and Chen, Jing and Lin, Jinghao and Chen, Jinhao and Wang, Jinjiang and Chen, Junjie and Lei, Leqi and Gong, Letian and Pan, Leyi and Liu, Mingdao and Xu, Mingde and Zhang, Mingzhi and Zheng, Qinkai and Lyu, Ruiliang and Tu, Shangqin and Yang, Sheng and Meng, Shengbiao and Zhong, Shi and Huang, Shiyu and Zhao, Shuyuan and Xue, Siyan and Zhang, Tianshu and Luo, Tianwei and Hao, Tianxiang and Tong, Tianyu and Jia, Wei and Li, Wenkai and Liu, Xiao and Zhang, Xiaohan and Lyu, Xin and Zhang, Xinyu and Fan, Xinyue and Huang, Xuancheng and Xue, Yadong and Wang, Yanfeng and Wang, Yanling and Wang, Yanzi and An, Yifan and Du, Yifan and Huang, Yiheng and Niu, Yilin and Shi, Yiming and Wang, Yu and Wang, Yuan and Yue, Yuanchang and Li, Yuchen and Liu, Yusen and Zhang, Yutao and Wang, Yuting and Zhang, Yuxuan and Xue, Zhao and Du, Zhengxiao and Hou, Zhenyu and Wang, Zihan and Zhang, Peng and Liu, Debing and Xu, Bin and Li, Juanzi and Huang, Minlie and Dong, Yuxiao and Tang, Jie},
  title = {{GLM-4.5V} and {GLM-4.1V-Thinking}: Towards Versatile Multimodal Reasoning with Scalable Reinforcement Learning},
  year = {2026},
  eprint = {2507.01006},
  archivePrefix = {arXiv},
  primaryClass = {cs.CV}
}

@inproceedings{wang2020cspnet,
  author = {Wang, Chien-Yao and Liao, Hong-Yuan Mark and Wu, Yueh-Hua and Chen, Ping-Yang and Hsieh, Jun-Wei and Yeh, I-Hau},
  title = {{CSPNet}: A New Backbone That Can Enhance Learning Capability of {CNN}},
  booktitle = {Proceedings of the IEEE/CVF Conference on Computer Vision and Pattern Recognition (CVPR) Workshops},
  year = {2020},
  pages = {390--391}
}

@article{wang_survey_2024,
	title = {A survey on large language model based autonomous agents},
	volume = {18},
	issn = {2095-2236},
	url = {https://doi.org/10.1007/s11704-024-40231-1},
	doi = {10.1007/s11704-024-40231-1},
	number = {6},
	journal = {Frontiers of Computer Science},
	author = {Wang, Lei and Ma, Chen and Feng, Xueyang and Zhang, Zeyu and Yang, Hao and Zhang, Jingsen and Chen, Zhiyuan and Tang, Jiakai and Chen, Xu and Lin, Yankai and Zhao, Wayne Xin and Wei, Zhewei and Wen, Jirong},
	month = mar,
	year = {2024},
	pages = {186345},
}

@inproceedings{wang2020minilm,
  author = {Wang, Wenhui and Wei, Furu and Dong, Li and Bao, Hangbo and Yang, Nan and Zhou, Ming},
  title = {{MiniLM}: Deep Self-Attention Distillation for Task-Agnostic Compression of Pre-Trained Transformers},
  booktitle = {Advances in Neural Information Processing Systems},
  volume = {33},
  pages = {5776--5788},
  year = {2020}
}

@inproceedings{xie2021segformer,
  author = {Xie, Enze and Wang, Wenhai and Yu, Zhiding and Anandkumar, Anima and Alvarez, Jose M. and Luo, Ping},
  title = {{SegFormer}: Simple and Efficient Design for Semantic Segmentation with Transformers},
  booktitle = {Advances in Neural Information Processing Systems},
  volume = {34},
  pages = {12077--12090},
  year = {2021}
}

@misc{xu2024rs,
  author = {Xu, Wenjia and Yu, Zijian and Mu, Boyang and Wei, Zhiwei and Zhang, Yuanben and Li, Guangzuo and Wang, Jiuniu and Peng, Mugen},
  title = {{RS-Agent}: Automating Remote Sensing Tasks through Intelligent Agent},
  year = {2026},
  eprint = {2406.07089},
  archivePrefix = {arXiv},
  primaryClass = {cs.CV}
}

@article{xu2024cas,
  author = {Xu, Yulin and Ouyang, Chaojun and Xu, Qingsong and Wang, Dongpo and Zhao, Bo and Luo, Yutao},
  title = {{CAS} Landslide Dataset: A Large-Scale and Multisensor Dataset for Deep Learning-Based Landslide Detection},
  journal = {Scientific Data},
  volume = {11},
  number = {1},
  pages = {12},
  year = {2024},
  doi = {10.1038/s41597-023-02847-z}
}

@article{yang_feature_2025,
	title = {A feature fusion method on landslide identification in remote sensing with {Segment} {Anything} {Model}},
	volume = {22},
	issn = {1612-5118},
	url = {https://doi.org/10.1007/s10346-024-02390-x},
	doi = {10.1007/s10346-024-02390-x},
	number = {2},
	journal = {Landslides},
	author = {Yang, Chuan and Zhu, Yueqin and Zhang, Jiantong and Wei, Xiaoqiang and Zhu, Haomeng and Zhu, Zhehui},
	month = feb,
	year = {2025},
	pages = {471--483},
}

@misc{yao2022react,
  author = {Yao, Shunyu and Zhao, Jeffrey and Yu, Dian and Du, Nan and Shafran, Izhak and Narasimhan, Karthik R. and Cao, Yuan},
  title = {{ReAct}: Synergizing Reasoning and Acting in Language Models},
  year = {2023},
  eprint = {2210.03629},
  archivePrefix = {arXiv},
  primaryClass = {cs.CL}
}

@inproceedings{yuan2020object,
  author = {Yuan, Yuhui and Chen, Xilin and Wang, Jingdong},
  title = {Object-Contextual Representations for Semantic Segmentation},
  booktitle = {Computer Vision -- ECCV 2020},
  publisher = {Springer International Publishing},
  address = {Cham},
  pages = {173--190},
  year = {2020}
}

@article{zhang_analysis_2024,
	title = {Analysis of the impact of terrain factors and data fusion methods on uncertainty in intelligent landslide detection},
	volume = {21},
	issn = {1612-5118},
	url = {https://doi.org/10.1007/s10346-024-02260-6},
	doi = {10.1007/s10346-024-02260-6},
	number = {8},
	journal = {Landslides},
	author = {Zhang, Rui and Lv, Jichao and Yang, Yunjie and Wang, Tianyu and Liu, Guoxiang},
	month = aug,
	year = {2024},
	pages = {1849--1864},
}

@inproceedings{zhao2017pyramid,
  author = {Zhao, Hengshuang and Shi, Jianping and Qi, Xiaojuan and Wang, Xiaogang and Jia, Jiaya},
  title = {Pyramid Scene Parsing Network},
  booktitle = {Proceedings of the IEEE Conference on Computer Vision and Pattern Recognition (CVPR)},
  year = {2017},
  pages = {2881--2890}
}

@inproceedings{zheng2024llamafactory,
  author = {Zheng, Yaowei and Zhang, Richong and Zhang, Junhao and Ye, Yanhan and Luo, Zheyan},
  title = {{LlamaFactory}: Unified Efficient Fine-Tuning of 100+ Language Models},
  booktitle = {Proceedings of the 62nd Annual Meeting of the Association for Computational Linguistics (Volume 3: System Demonstrations)},
  address = {Bangkok, Thailand},
  publisher = {Association for Computational Linguistics},
  pages = {400--410},
  year = {2024},
  doi = {10.18653/v1/2024.acl-demos.38}
}

@article{zhou2026landslide,
  author = {Zhou, Daoying and Liu, Huilin and Jin, Xiaowei and Wei, Qingjie and Cui, Kaiheng},
  title = {Landslide Detection in {UAV} Imagery: A Wavelet-Domain-Driven Multiscale Attention Approach},
  journal = {IEEE Transactions on Geoscience and Remote Sensing},
  volume = {64},
  pages = {1--15},
  year = {2026},
  doi = {10.1109/TGRS.2026.3660112}
}

\end{document}